\documentclass[10pt,a4paper]{article}
\usepackage[top=22mm,bottom=24mm,left=22mm,right=22mm]{geometry}
\usepackage{mathptmx}\usepackage[T1]{fontenc}\usepackage[utf8]{inputenc}
\usepackage{amsmath,amssymb,bm}\usepackage{graphicx}\usepackage{booktabs}
\usepackage{multirow}\usepackage{array}\usepackage{xcolor}
\usepackage{caption}\usepackage{subcaption}\usepackage{float}\usepackage{placeins}
\usepackage[hidelinks]{hyperref}
\floatstyle{ruled}\newfloat{algorithm}{tp}{loa}\floatname{algorithm}{Algorithm}
\graphicspath{{figures/}}
\newcommand{\SM}[1]{Supplementary Material~#1}
\newcolumntype{C}[1]{>{\centering\arraybackslash}p{#1}}
\newcolumntype{L}[1]{>{\raggedright\arraybackslash}p{#1}}

\hypersetup{pdftitle={REMBGO for Evaluation-Limited Optimization: CEC2022 Benchmarking and EAF-STATCOM Generalization},
pdfauthor={V. Chekuri, N. Ramarao, P. N. A.},
pdfsubject={Metaheuristic optimization; CEC2022 benchmarking; STATCOM control}}

\title{\bfseries REMBGO for Evaluation-Limited Optimization:\\
CEC2022 Benchmarking and EAF--STATCOM Generalization}
\author{Vikram Chekuri, Narapareddy Ramarao, Prashanth N A\\[3pt]
\small Department of Electrical and Electronics Engineering,\\
\small BMS Institute of Technology and Management, Bengaluru, Karnataka, India\\
\small Affiliated to Visvesvaraya Technological University, Belagavi, Karnataka,
India\\[2pt]
\small Corresponding author: vikramchekuri@bmsit.in}
\date{}
\begin{document}\maketitle

% =============================================================================
\begin{abstract}
Metaheuristic comparisons are often reported at evaluation budgets far larger
than an expensive engineering simulation can afford. This paper presents REMBGO,
an evaluation-budget-aware refinement of EMBGO. Its contribution is not any
single operator: it is the integration of population preservation,
scale-normalized exploration and recovery mechanisms whose parent-relative
benefit is concentrated under restricted function-evaluation budgets. REMBGO is
evaluated on CEC2022 at a primary, application-aligned budget of 1040 function
evaluations, with a surrounding budget ladder reported descriptively to
characterise budget dependence rather than as a family of controlled hypotheses.
A component-removal study finds no individual mechanism whose removal survives
multiplicity correction, consistent with an integrated rather than a
single-operator contribution. The algorithm is then applied to EAF--STATCOM
controller tuning, and the resulting frozen gain vectors are transferred without
retuning to twenty pre-specified unseen stochastic arc realizations ($210$
fixed-gain simulations, all feasible). The selected gain-vector main effect
accounts for only $3.6$--$5.8\%$ of the crossed validation variation, against
approximately $61.4\%$ for interaction plus residual, and the matched
REMBGO--EMBGO paired differences are controller-dependent in sign and not
statistically significant. REMBGO is therefore presented as a resource-aware
refinement whose benefit must be stated together with the evaluation budget,
not as a universally superior optimizer.
\end{abstract}

\vspace{4pt}

% =============================================================================
\section{Introduction}
Controller tuning by simulation can be computationally expensive because every
candidate gain vector requires a MATLAB/Simulink electromagnetic-transient (EMT)
simulation. In the EAF--STATCOM application considered here, the adopted tuning
protocol uses exactly $1040$ objective evaluations. This is much smaller than the largest
CEC2022~\cite{ref_cec22} budgets, so optimizer rankings obtained only at competition budgets
should not automatically be assumed to represent behaviour around $10^{3}$
evaluations.

Benchmark reports commonly emphasize final-quality performance at comparatively
large evaluation budgets, whereas the intended application here operates near
$10^{3}$ objective evaluations, and the ranking observed at a competition budget
need not preserve the same ordering when the methods are reconfigured to spend far
less. Schedules that anneal a search radius, a mutation
scale, or a step length over the run behave differently when the run is short, so
a method configured for a large budget and halted early is a different
configuration from the same method configured for a small budget. For an
application that can afford only about $10^{3}$ evaluations, the useful question
is how the methods compare at that budget when each budget is run as an
independently configured experiment rather than read from an early point of a
long run.

The optimizer studied here belongs to the multiplayer battle-game family. The
Multiplayer Battle Game Optimizer (MBGO)~\cite{ref_mbgo} models candidates as
players that improve through pairwise contests; the Efficient Multiplayer Battle
Game Optimizer (EMBGO)~\cite{ref_embgo} streamlines that scheme for numerical
optimization. REMBGO augments the EMBGO search structure with elitism, a
fitness-weighted centroid, normalized neighbourhood control, additional
perturbation operators, stagnation handling and partial restart, with the aim of
improving search when evaluations are limited, while retaining the battle-game
core. The study therefore addresses two linked questions: whether the REMBGO
refinements improve on EMBGO when objective evaluations are scarce, and whether
optimizer-selected controller gains retain their performance when exposed to
unseen stochastic EAF trajectories.

Budget-dependent benchmarking evaluates an optimizer at several
function-evaluation budgets rather than judging it only by its final result at one
large budget. This perspective, also referred to as anytime benchmarking, is
supported by the COCO framework~\cite{ref_coco} and its budget-dependent
extension~\cite{ref_tusar}. Restricted function-evaluation
budgets are a defining constraint of expensive optimization~\cite{ref_renmeng},
and methods aimed at that setting, such as emulation-based adaptive differential
evolution~\cite{ref_nishihara}, target the small-evaluation regime directly.
On the application side, metaheuristic STATCOM tuning has been reported with
hybrid ACO--PSO and differential-evolution/harmony-search
approaches~\cite{ref_kamel,ref_zhang}, broader VSC-based FACTS control is
reviewed in~\cite{ref_hussen}, and for electric-arc-furnace service, STATCOM
flicker mitigation has been addressed through conventional and energy-storage
STATCOMs~\cite{ref_virtanen}, nonlinear control~\cite{ref_yazdani}, and more
recently multi-objective evolutionary tuning~\cite{ref_xu}. The cited STATCOM and
EAF studies report performance at a fixed experimental setting and do not address
the budget-resolved and transfer questions examined here.

The methodological gap addressed here is the combined evaluation of a refined
optimizer across matched FE budgets, against its parent, at the exact application
budget, and on unseen stochastic application trajectories. This paper makes three
contributions.
\begin{enumerate}\itemsep1pt
\item A five-optimizer, thirty-run CEC2022 study over a common set of
independently configured evaluation budgets at $D{=}10$ and $D{=}20$, including
the exact $1040$-evaluation budget used in the application
(Sections~\ref{sec:setup}--\ref{sec:results}).
\item A parent-relative analysis of REMBGO against EMBGO, together with a
$1040$-evaluation component-removal analysis and an associated supplementary
sensitivity study.
\item A $210$-run fixed-gain EAF--STATCOM generalization study, in which one
selected gain vector per optimizer and controller is replayed across twenty
unseen stochastic arc realizations (Section~\ref{sec:results}).
\end{enumerate}
The comparison methodology follows Derrac et al.~\cite{ref_Derrac2011}, with Holm
correction~\cite{ref_holm} applied within each planned function family.

% =============================================================================
\section{Application Context and Objective Function}\label{sec:objective}
This section gives the application detail needed to read the fixed-gain transfer
study of Section~\ref{sec:results}; the paper is otherwise self-contained. A
STATCOM compensates a point of common coupling (PCC) that feeds an electric arc
furnace (EAF) whose time-varying nonlinear arc dynamics produce rapidly varying
reactive power and PCC-voltage modulation. The application uses an 11~kV, 50~Hz
EAF--STATCOM EMT plant with 5~MVA grid short-circuit capacity and $X/R=7$; a
500~kVA two-level VSC-STATCOM is connected in shunt at the PCC through its coupling
transformer and LCL filter. Two controllers are considered. The \emph{DQ-PI} controller uses decoupled current PI
control with filtered dc-link regulation, split-frequency reactive-power control,
bounded slow integral action, current limiting and anti-windup. The
\emph{Lyapunov} controller is a nonlinear design with full reactive-power PI
regulation, voltage droop, dc-link regulation and nonlinear inner current
feedback with the associated limiting. Each controller exposes six tunable
parameters,
\begin{equation}
\bm{g}_\mathrm{DQPI}=[K_{p,id},K_{i,id},K_{p,iq},K_{i,iq},K_{p,Q},T_{i,Q}],
\qquad
\bm{g}_\mathrm{Lyap}=[K_\mathrm{droop},K_{q,p},K_{q,i},k_{1,i},k_{i,i},k_{2,i}],
\label{eq:decvars}
\end{equation}
where for DQ-PI the reactive integral gain follows $K_{i,Q}=K_{p,Q}/T_{i,Q}$
under the parameter transformation used during optimization and the remaining
structure is fixed. The five optimizers (PSO, GWO, FP-SHADE, EMBGO, REMBGO) tune
each controller. Every run uses population $N=40$ and $25$ update iterations, so, including
initialization, each run spends $N(25+1)=1040$ objective evaluations
(Section~\ref{sec:formulation}); this is the exact application-aligned budget
carried into the benchmark study. During tuning the arc-noise stream is held at a
single stochastic realization, so repeated optimizer seeds measure optimizer
stochasticity rather than different furnace trajectories, a distinction central to
the transfer experiment.

The $1040$-evaluation limit is set by the cost of the objective, not by an
arbitrary benchmark choice. In the archived EAF--STATCOM tuning campaign, one
complete $1040$-evaluation optimization seed required approximately $98$--$273$ min
of wall-clock time depending on controller, worker count and computing platform.
Among the fourteen archived cases executed with eight workers, runtimes were approximately
$98$--$131$ min per seed. These measurements are reported only to establish the
practical cost of objective evaluations and are not used to compare optimizer
execution speed. The twenty archived optimizer--scenario combinations correspond to
five optimizers evaluated across four controller--EAF scenarios (two controllers
$\times$ two EAF conditions).

Each candidate is scored by an EMT simulation over the post-transient window
$t\in[0.50,1.00]$~s. The performance term combines a reactive-power fluctuation
feature and a one-cycle-RMS PCC-voltage feature, each normalized by a common
uncompensated deterministic reference so that the score measures disturbance
severity consistently,
\begin{equation}
T_Q=\frac{Q_\mathrm{fluc}}{Q_\mathrm{fluc}^{\mathrm{ref}}},\quad
T_V=\frac{V_\mathrm{pcc}}{V_\mathrm{pcc}^{\mathrm{ref}}},\qquad
J_\mathrm{perf}=0.60\,T_Q+0.40\,T_V,
\label{eq:jperf}
\end{equation}
and the scored scalar adds graded quality and safety penalties,
\begin{equation}
J_\mathrm{tot}=J_\mathrm{perf}+P_{\overline Q}+P_{V_\mathrm{dc}}+P_\mathrm{safety}.
\label{eq:jtot}
\end{equation}
Here $J_\mathrm{tot}$ is a short-window optimization score rather than an IEC
flicker index. $P_{\overline Q}$ and $P_{V_\mathrm{dc}}$ are graded penalties on mean
reactive-power draw beyond a small allowance and on dc-link peak deviation beyond
a $3\%$ band, and $P_\mathrm{safety}$ combines hard feasibility gates---triggered
when mean or peak reactive power exceeds the compensator rating or the dc-link
voltage leaves its admissible envelope---with graded soft costs on converter
over-modulation and current-tracking error. A candidate is \emph{hard-feasible}
when no hard gate is triggered. A compact self-contained statement of the feature
configuration and penalty thresholds is given in \SM{S5}. The single scalar
$J_\mathrm{tot}$ is the only objective used in both the application and the
transfer study.

% =============================================================================
\section{REMBGO Formulation}\label{sec:formulation}
This section specifies REMBGO in enough detail to reimplement it. The algorithm
defined here is the one used in all reported experiments.

\subsection{Population model and notation}
REMBGO evolves $\mathbf{X}=\{\bm{x}_1,\dots,\bm{x}_N\}$, $\bm{x}_i\in
\mathbb{R}^{D}$, within bounds $\bm{l},\bm{u}$ and range
$\bm{r}=\bm{u}-\bm{l}$. The run spends a fixed objective-evaluation budget $B$ over
$T_B=\lceil(B-N)/N\rceil$ generations; at generation $t$ the normalized progress is
$\tau=t/T_B\in(0,1]$. The population is
sorted so $f_1\le\dots\le f_N$; $\bm{x}_b=\bm{x}_1$, $\bm{x}_w=\bm{x}_N$.
Normalized coordinates are $\hat{\bm{x}}=(\bm{x}-\bm{l})\oslash\bm{r}$, and
diversity is
\begin{equation}
\mathrm{div}=\tfrac{1}{D}\textstyle\sum_{d=1}^{D}
\operatorname{std}_i(\hat x_{i,d}).
\label{eq:div}
\end{equation}

\subsection{Inherited EMBGO structure and its limitations}
EMBGO~\cite{ref_embgo} advances each player through one of three search
behaviours. A player inside a safe region around the current best moves by a
current-to-best-and-mean differential update (sinusoidally weighted differential
vectors toward the best and the population mean); a
player outside that region performs L\'evy-flight exploration; and in a
separate opponent battle a random opponent sets the direction, the player
recombining with a stronger opponent by dimension-wise crossover and otherwise
taking a single cosine-modulated step. Three features limit this
scheme for a simulation-expensive, penalty-constrained objective:
the arithmetic-mean attractor is pulled toward poor or penalised candidates; the
neighbourhood radius is measured in raw coordinate units and is
dominated by the widest-ranging variable; and although greedy
replacement protects a parent from a worse trial, EMBGO keeps no explicit
elite set and no stagnation or diversity recovery, so search
effort is spent perturbing even the current best and a prematurely converged
population is never re-diversified.

\subsection{REMBGO search cycle}
REMBGO retains the battle-game core and addresses these three features. Before the
equations, the search cycle is read as a fixed sequence. Each generation
(i)~ranks the current population and identifies the best and worst players;
(ii)~preserves a few elite players unchanged; (iii)~measures population diversity
and forms a fitness-weighted centroid, an adaptive neighbourhood radius, a decaying
step scale and an adaptive patience; (iv)~generates one trial for every non-elite
player by selecting a single operator from a fixed portfolio; (v)~applies a
stagnation-dependent jitter and repairs any boundary violation; (vi)~evaluates the
trials while respecting the remaining evaluation budget and keeps, for each player,
the better of parent and trial; and (vii)~detects stagnation or low diversity and
partially restarts the weakest players. The cycle repeats until the
function-evaluation budget is spent.

The components are specified below as R1--R11 under three functional headings; the
labels are retained for cross-reference, and each component is marked as inherited
from EMBGO, inherited but modified, or added in REMBGO. Notation follows the
implementation, except that the centroid-eligible index set is written
$\mathcal{I}_{\mathrm{ctr}}$, to avoid confusion with hard-feasibility in the
application, and the neighbourhood radius as $R_t$, to make its iteration
dependence explicit.

\subsubsection*{Population guidance and preservation}
\textbf{(R1) Rank-based elitism [added].} After ranking the population, the
$K_e{=}3$ best players are excluded from trial generation and copied unchanged
into the corresponding trial rows; their stored fitness values are restored after
the generation-level evaluation. These top-ranked states are thus preserved while
the remaining $N-K_e$ players undergo search updates, an explicit preservation
distinct from EMBGO's inherited greedy replacement.

\textbf{(R2) Fitness-weighted centroid [added].} A centroid attractor is formed
over the centroid-eligible players
$\mathcal{I}_{\mathrm{ctr}}=\{i:f_i<f_{\mathrm{ctr}}\}$, where
$f_{\mathrm{ctr}}=10^{7}$ is a fixed eligibility constant that screens out
high-fitness candidates, and
$f_{\min}=\min_{j\in\mathcal{I}_{\mathrm{ctr}}}f_j$,
\begin{equation}
w_i=\frac{1/(f_i-f_{\min}+\varepsilon)}
{\sum_{j\in\mathcal{I}_{\mathrm{ctr}}}1/(f_j-f_{\min}+\varepsilon)},\qquad
\bm{c}=\sum_{i\in\mathcal{I}_{\mathrm{ctr}}}w_i\bm{x}_i,
\label{eq:centroid}
\end{equation}
with $\varepsilon=10^{-6}$. Candidates above the eligibility threshold are
excluded, while poorer eligible candidates receive smaller centroid weights; if
$|\mathcal{I}_{\mathrm{ctr}}|<3$, the plain population mean
\[
\bm{c}=\frac{1}{N}\sum_{i=1}^{N}\bm{x}_i
\]
is used.

\textbf{(R3) Normalized decaying neighbourhood radius [modified].} The raw-unit
EMBGO radius is replaced by one measured in normalized coordinates and shrunk over
the run,
\begin{equation}
\delta\sim\mathcal U(0.8,1.2),\quad \rho(\tau)=1.5-0.7\tau,\quad
R_t=\lVert\hat{\bm x}_b-\hat{\bm x}_w+\varepsilon_R\bm 1\rVert_2\,\delta\,\rho(\tau),
\quad \varepsilon_R=10^{-12},
\label{eq:radius}
\end{equation}
where the scalar $\varepsilon_R$ is added componentwise inside the norm. One value
of $\delta$ is sampled at the start of each generation and is shared by all
neighbourhood tests in that generation.

\subsubsection*{Trial-generation portfolio}
\textbf{(R4) Operator selection [modified].} Each non-elite player tests fresh
independent uniform draws sequentially, and a later draw is made only when all
preceding branch tests fail: opposition-based learning~\cite{ref_obl} if
$u_1<p_o=0.10$ [added]; otherwise best-centred Cauchy mutation if $u_2<p_c=0.20$
[added]; otherwise the radius-gated battle-game core if $u_3<0.5$; otherwise the
opponent battle. Because the tests are sequential, the realized branch
probabilities are
\begin{equation}
\begin{aligned}
P_\mathrm{OBL}&=0.10, & P_\mathrm{Cauchy}&=0.90\times0.20=0.18,\\
P_\mathrm{core}&=0.90\times0.80\times0.50=0.36, &
P_\mathrm{battle}&=0.36 .
\end{aligned}
\label{eq:branchprob}
\end{equation}
The opposition move reflects the player about a scaled box centre,
\begin{equation}
\beta\sim\mathcal U(0.3,0.7),\qquad
\bm{x}'=\bm{x}_i+\beta(\bm{l}+\bm{u}-2\bm{x}_i),
\label{eq:obl}
\end{equation}
and the Cauchy move is centred on the incumbent best rather than on the player,
\begin{equation}
\bm{s}=\lambda(\tau)\,\bm{r}\,(1-0.5\tau),\qquad
\bm{x}'=\bm{x}_b+\bm{s}\odot\tan\!\bigl[\pi(\bm{U}-\tfrac12)\bigr],
\quad \bm{U}\sim\mathcal U(0,1)^D ,
\label{eq:cauchy}
\end{equation}
following recent use of Cauchy perturbation for numerical
optimization~\cite{ref_cauchy}. The player selected for a Cauchy trial supplies
only the parent fitness for the later greedy test.

\textbf{(R5) Step schedule [added].} The common base gain $\lambda(\tau)=
\lambda_{\max}-(\lambda_{\max}-\lambda_{\min})\tau$ decreases linearly from
$\lambda_{\max}=0.02$ to $\lambda_{\min}=0.003$. The L\'evy move uses this gain
directly, whereas the Cauchy move applies the additional attenuation factor
$1-0.5\tau$ of~\eqref{eq:cauchy}.

\textbf{(R6) Radius-gated core [inherited core; added triple-leader].} A player
within the neighbourhood radius, $\lVert\hat{\bm x}_i-\hat{\bm x}_b\rVert\le R_t$,
takes a grey-wolf triple-leader blend~\cite{ref_mirjalili_gwo} with probability
$0.30$,
\begin{equation}
\bm{X}_k=\bm{x}_{(k)}-\bm{A}_k\odot|\bm{C}_k\odot\bm{x}_{(k)}-\bm{x}_i|,
\qquad \bm{x}'=\tfrac13(\bm{X}_1+\bm{X}_2+\bm{X}_3),
\label{eq:gwoblend}
\end{equation}
where $\bm{x}_{(1)},\bm{x}_{(2)},\bm{x}_{(3)}$ are the first-, second- and
third-ranked players, $a=2(1-\tau)$, and each leader uses an independent
coefficient pair
$\bm{A}_k=2a\bm{r}_{A,k}-a$, $\bm{C}_k=2\bm{r}_{C,k}$ with
$\bm{r}_{A,k},\bm{r}_{C,k}\overset{\mathrm{iid}}{\sim}\mathcal U(0,1)^D$,
$k=1,2,3$; otherwise it takes the current-to-best-and-centroid update inherited
from EMBGO~\cite{ref_embgo}, realised as sinusoidally weighted differential
vectors,
\begin{equation}
\bm{x}'=\bm{x}_i+(\bm{x}_b-\bm{x}_i)\sin(2\pi r_1)
+(\bm{c}-\bm{x}_i)\sin(2\pi r_2) .
\label{eq:sinpull}
\end{equation}

\textbf{(R7) L\'evy escape [inherited form; modified step schedule].} A player
outside the radius explores by a L\'evy flight, the escape behaviour inherited
from EMBGO~\cite{ref_embgo} under REMBGO's decaying step schedule,
\begin{equation}
\bm{x}'=\bm{x}_i+\lambda(\tau)\,\bm{r}\odot
\frac{\bm{u}_L}{|\bm{v}_L|^{1/\beta_L}+\varepsilon_R},\qquad
\bm{u}_L\sim\mathcal N(0,\sigma_L^2),\ \bm{v}_L\sim\mathcal N(0,1),
\label{eq:levy}
\end{equation}
with $\beta_L=1.5$ and $\sigma_L$ the Mantegna scale~\cite{ref_mantegna}. The
regularizer $\varepsilon_R=10^{-12}$ used in~\eqref{eq:radius}
and~\eqref{eq:levy} keeps a collapsed population or a near-zero Gaussian draw from
producing a non-finite step and is numerically negligible at the scales
encountered here.

\textbf{(R8) Opponent battle [inherited].} Following the EMBGO battle
mechanism~\cite{ref_embgo}, an opponent $e\neq i$ is drawn
uniformly, and the contest direction depends on relative fitness,
\begin{equation}
\bm{d}=\begin{cases}\bm{x}_i-\bm{x}_e, & f_i\le f_e,\\
\bm{x}_e-\bm{x}_i, & f_i>f_e.\end{cases}
\label{eq:battledir}
\end{equation}
When the opponent is strictly better ($f_e<f_i$) the player recombines with it by
a dimension-wise crossover with fresh $r\sim\mathcal U(0,1)$ per coordinate,
\begin{equation}
x'_k=\begin{cases}x_{i,k}+d_k\,r, & r<0.5,\\
x_{e,k}+d_k\,r, & r\ge0.5;\end{cases}
\label{eq:crossover}
\end{equation}
otherwise it takes a single cosine-modulated step,
$\bm{x}'=\bm{x}_i+\bm{d}\cos(2\pi r)$.

\subsubsection*{Recovery and survivor handling}
\textbf{(R9) Stagnation jitter [added].} With stagnation counter $c_s$ and
patience $P$, a Gaussian jitter is added to each trial with a probability that
grows as the search stalls,
\begin{equation}
p_j=j_0+(j_{\max}-j_0)\min\!\Bigl(1,\frac{c_s}{P}\Bigr),\qquad
\bm{x}'\mathrel{+}=\eta\,\bm{r}\odot\mathcal N(0,1),
\label{eq:jitter}
\end{equation}
$j_0=0.05$, $j_{\max}=0.25$, $\eta=0.015$.

\textbf{(R10) Reflective repair [added] and greedy selection [inherited].} Each
coordinate outside $[\bm l,\bm u]$ is reflected back into range,
$x_k\leftarrow 2l_k-x_k$ or $2u_k-x_k$, up to ten times, then clamped. The trial
replaces its parent only when strictly better:
$\bm{x}_i^{t+1}=\bm{x}'$ if $f(\bm{x}')<f_i$, else $\bm{x}_i^{t+1}=\bm{x}_i$.

\textbf{(R11) Adaptive patience and partial restart [added].} Patience follows
diversity,
\begin{equation}
P=\begin{cases}P_0+1, & \mathrm{div}>0.30,\\
\max(1,P_0-1), & \mathrm{div}<0.05,\\ P_0, & \text{otherwise},\end{cases}
\qquad P_0=3 ,
\label{eq:patience}
\end{equation}
and a restart fires when $c_s\ge P$, or when $\mathrm{div}<0.05$ and $c_s>1$. It
regenerates the worst $n_r=\operatorname{round}(0.30N)$ players in three groups:
uniform random draws in $[\bm l,\bm u]$; multiplicative neighbours of a reference
point $\bm{x}_\mathrm{ref}\odot(1+0.3\,\mathcal U[-1,1]^D)$; and Gaussian
neighbours of the incumbent $\bm{x}_b+0.1\,\bm{r}\odot\mathcal N(0,1)$. At $N=40$
this gives $n_r=12$ with an executed split of $6/4/2$. Restarted positions are
inserted with stored fitness $f=+\infty$, so they enter the next generation as
unevaluated parent states and the next finite evaluated trial at those indices is
accepted by the strict greedy rule; the counter then resets to $c_s=0$. The restart
targets the worst $n_r$ post-selection players, which under the reported
finite-fitness runs excludes the current elite set. On the symmetric CEC2022
bounds the reference point
defaults to the box midpoint (the origin), so the multiplicative group returns
identical reference copies; the reported results are conditional on this
implemented behaviour, and a matched diagnostic using an alternative
non-degenerate reference is documented in \SM{S3}.

\subsection{Evaluation accounting}

Elite handling carries a stated cost. Each generation copies the $K_e{=}3$ elites
into the trial population and generates only $N-K_e=37$ new trials, yet the
implementation evaluates all $N=40$ trials and restores the stored elite fitnesses
before selection, so one generation spends $40$ evaluations for $37$ new
candidates and a run of $T$ generations costs $N(T+1)$ evaluations including
initialization. When $B$ is not an integer multiple of $N$ beyond initialization
the final generation is evaluated only partially (Algorithm~\ref{alg:rembgo}), so
the requested budget is met exactly. Every algorithm receives the same budget with
objective calls counted identically, so the comparison is matched in objective-call
budget rather than in newly generated candidates: REMBGO spends three of every
forty generation-level calls re-evaluating copied elite rows ($3/40=7.5\%$), which
the other four methods do not, amounting at the $1040$-evaluation budget to
$25\times3=75$ of $1040$ calls ($7.21\%$); this overhead is retained without
compensation. Excluding objective evaluation, one REMBGO generation requires
$O(ND)$ arithmetic for population statistics, trial generation and boundary repair
together with $O(N\log N)$ sorting, with $O(ND)$ population storage. In the intended
EAF--STATCOM application this internal overhead is negligible relative to the EMT
objective evaluations.
Algorithm~\ref{alg:rembgo} states the accounting explicitly, and
Table~\ref{tab:rembgo_defaults} lists the default control parameters.

\begin{algorithm}[t]
\caption{REMBGO with strict function-evaluation accounting.}\label{alg:rembgo}
\footnotesize
\begin{flushleft}\setlength{\parindent}{0pt}
\textbf{Input:} bounds $\bm l,\bm u$; population size $N$; evaluation budget $B$;
elite count $K_e$; restart reference point $\bm{x}_\mathrm{ref}$ (box midpoint on
the CEC benchmark, not inserted into the initial population).\\
\textbf{Output:} best solution $\bm{x}_b$ and its value $f_b$.\\[3pt]
1:\ $T_B\leftarrow\lceil(B-N)/N\rceil$.\\
2:\ Initialize the $N$ candidates and evaluate them; $\mathrm{fe}\leftarrow N$,
$c_s\leftarrow0$; store the global best $(\bm x_b,f_b)$.\\
3:\ \textbf{for} $t=1,\dots,T_B$ \textbf{do}\\
4:\ \hspace{1em}$\tau\leftarrow t/T_B$.\\
5:\ \hspace{1em}\emph{Rank and preserve.} Sort by fitness; set best $\bm x_b$ and
worst $\bm x_w$; compute $\mathrm{div}$~\eqref{eq:div} and patience
$P$~\eqref{eq:patience}; copy the $K_e$ elites unchanged into the trial rows
(R1).\\
6:\ \hspace{1em}\emph{Prepare guidance.} Compute centroid $\bm c$~\eqref{eq:centroid},
sample $\delta$ and set radius $R_t$~\eqref{eq:radius}, step scale
$\lambda(\tau)$ and jitter probability $p_j$~\eqref{eq:jitter}.\\
7:\ \hspace{1em}\textbf{for} each non-elite player $i$ \textbf{do}\\
8:\ \hspace{2em}select the first active operator (R4): opposition~\eqref{eq:obl}
$\rightarrow$ Cauchy~\eqref{eq:cauchy} $\rightarrow$ radius-gated core
(triple-leader~\eqref{eq:gwoblend} or centroid update~\eqref{eq:sinpull} if
$\lVert\hat{\bm x}_i-\hat{\bm x}_b\rVert\le R_t$, else L\'evy~\eqref{eq:levy})
$\rightarrow$ opponent battle~\eqref{eq:battledir}--\eqref{eq:crossover}; form the
trial.\\
9:\ \hspace{2em}apply stagnation jitter with probability $p_j$ (R9); repair
boundary violations by reflection and clamping (R10).\\
10:\ \hspace{1em}\textbf{end for}\\
11:\ \hspace{1em}\emph{Budgeted evaluation.}
$m\leftarrow\min(N,\,B-\mathrm{fe})$; evaluate the first $m$ trial rows;
$\mathrm{fe}\leftarrow\mathrm{fe}+m$; unevaluated trials keep their parents;
restore the stored elite fitnesses.\\
12:\ \hspace{1em}\emph{Survivor selection.} Apply strict one-to-one greedy
parent--trial selection (R10); update $(\bm x_b,f_b)$; set $c_s\leftarrow0$ if the
best improved, else $c_s\leftarrow c_s+1$.\\
13:\ \hspace{1em}\emph{Recovery.} \textbf{if} $c_s\ge P$ \textbf{or}
($\mathrm{div}<0.05$ \textbf{and} $c_s>1$) \textbf{then} regenerate the worst
$n_r=\operatorname{round}(0.30N)$ players by the $6/4/2$ construction, set their
stored fitness to $+\infty$, and reset $c_s\leftarrow0$ (R11).\\
14:\ \hspace{1em}\textbf{if} $\mathrm{fe}\ge B$ \textbf{then break}.\\
15:\ \textbf{end for}\\
16:\ \textbf{return} $\bm{x}_b,f_b$.
\end{flushleft}
\end{algorithm}

\begin{table}[t]\centering
\caption{Default REMBGO control parameters. The Cauchy and triple-leader
probabilities, elite count and restart fraction are varied only in the sensitivity
analysis of \SM{S2}.}
\label{tab:rembgo_defaults}
\setlength\tabcolsep{5pt}\renewcommand{\arraystretch}{1.1}\footnotesize
\begin{tabular}{L{6.0cm}C{2.0cm}c}
\toprule
\textbf{Parameter} & \textbf{Symbol} & \textbf{Value}\\ \midrule
Elite count & $K_e$ & 3\\
Radius scale & $\delta$ & $\mathcal U(0.8,1.2)$\\
Radius decay & $\rho(\tau)$ & $1.5\rightarrow0.8$\\
Step gain & $\lambda$ & $0.02\rightarrow0.003$\\
L\'evy exponent & $\beta_L$ & 1.5\\
Opposition probability & $p_o$ & 0.10\\
Cauchy probability (conditional) & $p_c$ & 0.20\\
Triple-leader probability & $p_T$ & 0.30\\
Jitter probability & $j_0\rightarrow j_{\max}$ & $0.05\rightarrow0.25$\\
Jitter scale & $\eta$ & 0.015\\
Base patience & $P_0$ & 3 (range 2--4)\\
Restart fraction & $r_f$ & 0.30 ($n_r=12$ at $N{=}40$)\\
Diversity thresholds & --- & 0.05, 0.30\\ \bottomrule
\end{tabular}
\end{table}

% =============================================================================
\FloatBarrier
\section{Experimental Design and Statistical Methods}\label{sec:setup}
The benchmarking protocol follows published methodological guidance for comparative
bio-inspired optimization studies~\cite{ref_latorre3}: a matched evaluation budget,
shared initialization, repeated runs and nonparametric paired testing.
\subsection{Algorithms}
PSO~\cite{ref_kennedy_pso} uses inertia $0.9\rightarrow0.4$, $c_1=c_2=2$ and a
velocity clamp of $0.2\bm{r}$; GWO~\cite{ref_mirjalili_gwo} uses
$a:2\rightarrow0$. FP-SHADE is a fixed-population implementation of
success-history adaptive differential evolution using current-to-$p$best/1
mutation, an external archive and success-history adaptation of $F$ and $CR$
($H=6$, $p=0.1$, archive rate $1.0$); linear population-size reduction is not
used~\cite{ref_shade}. Results for FP-SHADE are therefore not evidence about
population-reducing SHADE variants. All five implementations share a strict
evaluation-budget controller and per-evaluation best-so-far recording.

\subsection{Benchmark and evaluation budgets}
The CEC2022 suite~\cite{ref_cec22} provides twelve functions on $[-100,100]^D$:
F1 unimodal, F2--F5 basic multimodal, F6--F8 hybrid, F9--F12 composition, with
known optima $F^\ast$. The five optimizers are compared at $D\in\{10,20\}$ with
population $N=40$, thirty official runs per function, the official CEC2022
\texttt{Rand\_Seeds} list with the prescribed Mersenne-Twister generator, and the
identical uniform-random initial population $\mathbf X_0$ supplied to every
optimizer for each matched (function, dimension, run). No warm-start row is
inserted.

Each evaluation budget is run as a separate, independently configured experiment
rather than read as a checkpoint from a single long run. Because the safe radius,
mutation scale and step schedules are driven by the consumed fraction $\tau$ of
the iteration budget, a run configured for a large budget and halted early is a
different configuration from a method configured for the smaller budget. For a
requested budget $B$ the iteration count is
\begin{equation}
T_B=\Bigl\lceil\frac{B-N}{N}\Bigr\rceil ,
\label{eq:tbudget}
\end{equation}
and strict function-evaluation accounting is enforced. The common ladder spans
fourteen budgets per dimension:
\[
\text{$D{=}10$: }200,\,316,\,502,\,796,\,840,\,1040,\,1261,\,2000,\,3169,\,
5023,\,7962,\,12619,\,20000,\,200000\ \text{FEs};
\]
\[
\text{$D{=}20$: }125,\,227,\,414,\,754,\,840,\,1040,\,1373,\,2500,\,4551,\,
8286,\,15085,\,27464,\,50000,\,1000000\ \text{FEs},
\]
for a total of $5\times12\times30\times28=50{,}400$ optimizer runs. The largest
budget in each dimension matches the CEC2022 technical-report budget. The
$1040$-evaluation rung is one ordinary rung of this common ladder and is the exact
budget used in the EAF--STATCOM tuning application ($N=40$, $25$ updates,
$40\times(25+1)=1040$); it is referred to as the exact application-aligned budget.
The complete twenty-eight-rung ladder is tabulated in \SM{S1}.

\subsection{Statistical methods}
The terminal error of a run is $e=f_{\mathrm{best}}-F^{\ast}$, the gap between the
best objective found and the known optimum; runs with $e<10^{-8}$ are recorded as
$e=0$, and where a base-10 logarithm is taken the per-function median error is
floored at $10^{-8}$ (that is, $\log_{10}\max(e,10^{-8})$) so that the
$\Delta\log_{10}$ values of the component-removal study are well defined. Each
benchmark function is one statistical block. For every (dimension, budget)
condition the per-function summary is the median terminal error over the thirty
runs; the algorithms are ranked once per function, giving a rank matrix that
drives the Friedman test and the mean ranks. Because the algorithms share seeds
and identical initial populations, the primary pairwise test is the two-sided
exact Wilcoxon signed-rank test on the thirty matched pairs per function, with the
direction taken from the signed-rank statistic and Holm correction applied across
the twelve functions separately for each planned comparison~\cite{ref_holm}. Win/
tie/loss (W/T/L) counts report REMBGO better / no significant difference / EMBGO
better after this correction. The Holm family is the twelve functions within a
single budget; cross-rung multiplicity is not controlled, so budgets are not
compared inferentially against one another. The component-removal analysis of
Section~\ref{sec:results} uses the same matched CEC2022 records at $1040$
evaluations; its ranking, paired-comparison, familywise-correction and
exact-zero/tie conventions are specified in \SM{S5}.

\subsection{Reproducibility}
Each benchmark run is archived separately with its optimizer, dimension, function,
seed and evaluation-budget metadata. The reported tables and figures regenerate
directly from these stored records without rerunning the optimizers. For the
application transfer study, replay of the ten tuning-arc cases reproduces the
stored $J_\mathrm{tot}$ values to a maximum absolute difference of
$2.03\times10^{-8}$.

% =============================================================================
\FloatBarrier
\section{Results}\label{sec:results}

\subsection{Optimizer ranking at different evaluation budgets}
\begin{figure}[tbp]\centering
\includegraphics[width=\linewidth]{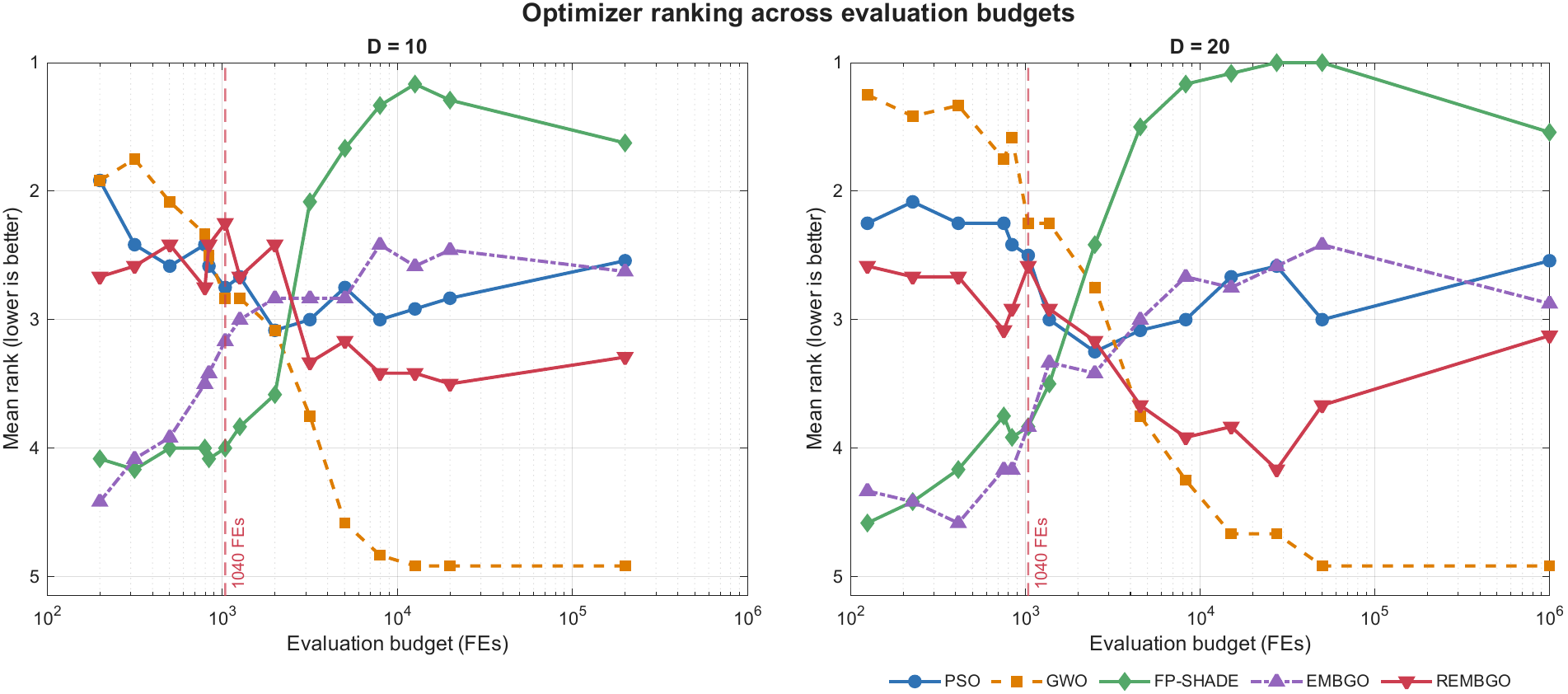}
\caption{Friedman mean rank of the five optimizers against the evaluation budget,
for $D{=}10$ (left) and $D{=}20$ (right); lower mean rank is better. The dashed
vertical marker identifies the exact $1040$-evaluation application budget.}
\label{fig:ranking}
\end{figure}

Figure~\ref{fig:ranking} shows the Friedman mean rank of each optimizer as a
function of the evaluation budget, and Table~\ref{tab:key_budget_results} lists
representative rungs at the two dimensions. The ranking changes materially with
the budget, and three regimes are visible.

At the smallest budgets the swarm methods lead: at $D{=}10$ with $200$
evaluations PSO and GWO share the lowest mean rank ($1.917$), ahead of
REMBGO ($2.667$), and at $D{=}20$ with $125$ evaluations GWO leads at $1.250$, so
REMBGO is competitive but not strongest at these very restricted budgets.

At the exact application-aligned budget of $1040$ evaluations the picture is
dimension-dependent. At $D{=}10$ REMBGO obtains the lowest observed mean rank
($2.250$, ahead of PSO $2.750$ and GWO $2.833$), but the omnibus test does not
reject ($p=0.089$), so REMBGO holds the lowest observed mean rank rather than being
the statistically strongest optimizer. At $D{=}20$ GWO leads ($2.250$), followed by
PSO ($2.500$) and REMBGO ($2.583$), yet REMBGO still ranks below its parent EMBGO
($3.833$) while the omnibus test rejects equal ranks ($p=0.022$); the planned EMBGO
contrast is reported next.

At large budgets FP-SHADE obtains the lowest observed mean ranks---$1.625$ at
$D{=}10$ with $200000$ evaluations and $1.542$ at $D{=}20$ with $10^{6}$---while
GWO falls to last in both. The ordering observed at very large CEC-style budgets is
therefore not preserved around the application-scale budget of $10^{3}$
evaluations.

\begin{table}[tb]\centering
\caption{Representative benchmark rungs across the three budget regimes; lower
mean rank is better and \emph{Best} names the optimizer with the lowest observed
mean rank. W/T/L reports REMBGO better\,/\,no significant difference\,/\,EMBGO
better over the twelve functions after within-budget Holm correction; the full
ladder is in \SM{S1}.}
\label{tab:key_budget_results}
\small
\setlength\tabcolsep{5pt}\renewcommand{\arraystretch}{1.12}
\begin{tabular}{rrrrrrrlcc}
\toprule
$D$ & FEs & PSO & GWO & FP-SHADE & EMBGO & REMBGO & Best & Friedman $p$ & W/T/L \\
\midrule
10 & 200    & 1.917 & 1.917 & 4.083 & 4.417 & 2.667 & PSO/GWO  & 1.93e-05 & 10/2/0 \\
10 & 1040   & 2.750 & 2.833 & 4.000 & 3.167 & 2.250 & REMBGO   & 0.0892   & 4/6/2  \\
10 & 200000 & 2.542 & 4.917 & 1.625 & 2.625 & 3.292 & FP-SHADE & 1.07e-06 & 0/9/3  \\
\midrule
20 & 125    & 2.250 & 1.250 & 4.583 & 4.333 & 2.583 & GWO      & 7.66e-08 & 10/2/0 \\
20 & 1040   & 2.500 & 2.250 & 3.833 & 3.833 & 2.583 & GWO      & 0.0224   & 9/1/2  \\
20 & 1000000& 2.542 & 4.917 & 1.542 & 2.875 & 3.125 & FP-SHADE & 3.82e-06 & 0/8/4  \\
\bottomrule
\end{tabular}
\end{table}

\subsection{REMBGO relative to its parent}
\begin{figure}[tbp]\centering
\includegraphics[width=\linewidth]{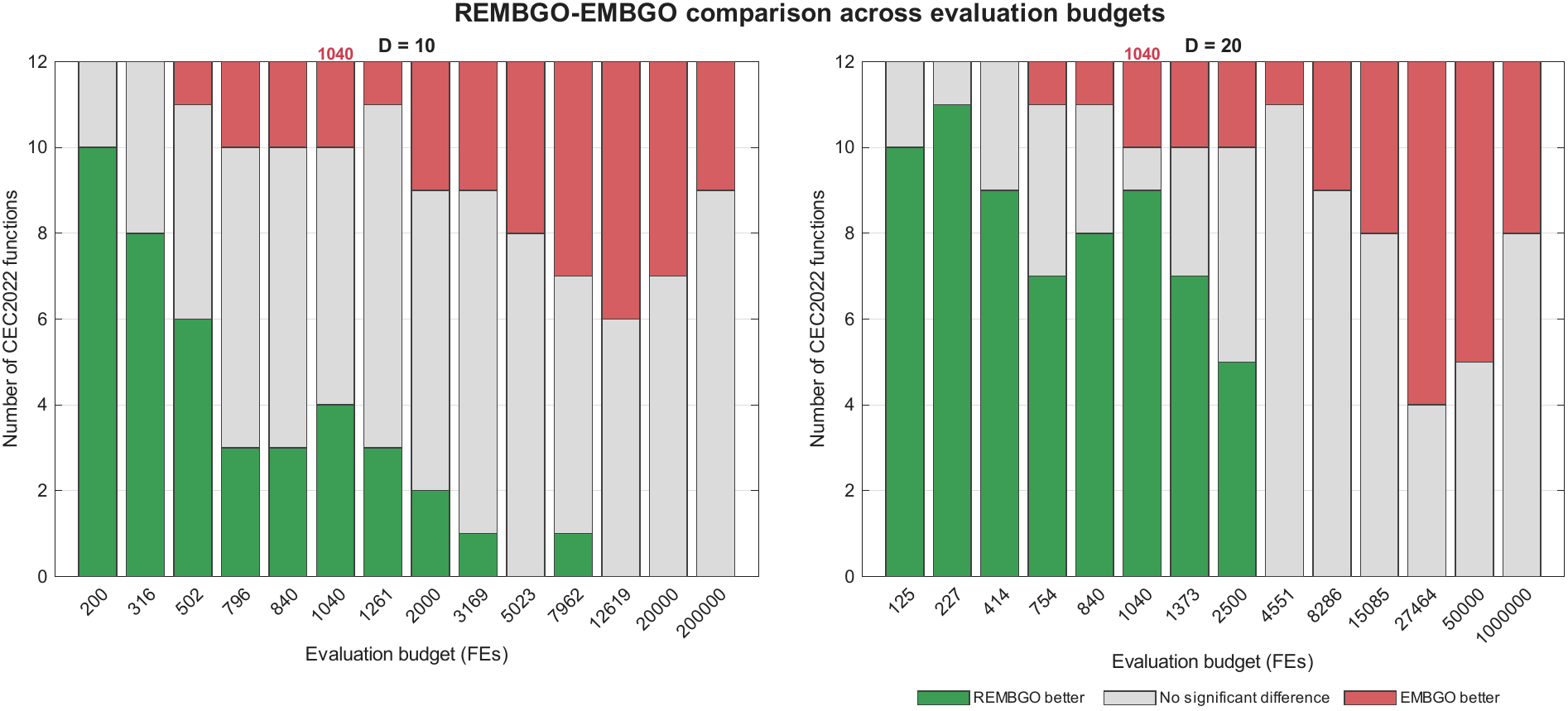}
\caption{Holm-corrected paired function-level comparison between REMBGO and EMBGO
across evaluation budgets. Green, grey and red mark functions significantly
favouring REMBGO, with no significant difference, and favouring EMBGO,
respectively; correction is applied across the twelve functions at each budget.}
\label{fig:parent}
\end{figure}

Because REMBGO is a refinement of EMBGO, the parent comparison is the most direct
test of the proposed modifications.
Figure~\ref{fig:parent} shows the Holm-corrected function-level record across the
budget ladder. At the exact application budget the record is $4/6/2$ at $D{=}10$
and $9/1/2$ at $D{=}20$; the $D{=}20$ case is notable because, even though GWO
holds the lowest overall mean rank there, REMBGO improves on its parent on nine of
the twelve functions after within-budget correction. The advantage is strongest
when evaluations are scarce---$10/2/0$ at $D{=}10$ with $200$ evaluations and
$11/1/0$ at $D{=}20$ with $227$---and narrows and reverses as the budget grows,
reaching $0/9/3$ at $D{=}10$ and $0/8/4$ at $D{=}20$ at the largest budgets. The
integrated modifications are therefore useful mainly in the evaluation-limited
regime and should not be presented as improving EMBGO at arbitrary budgets; a
precise crossover budget is not estimated, because cross-rung multiplicity is not
controlled.

\subsection{Component behaviour at the application budget}
\begin{figure}[tbp]\centering
\includegraphics[width=\linewidth]{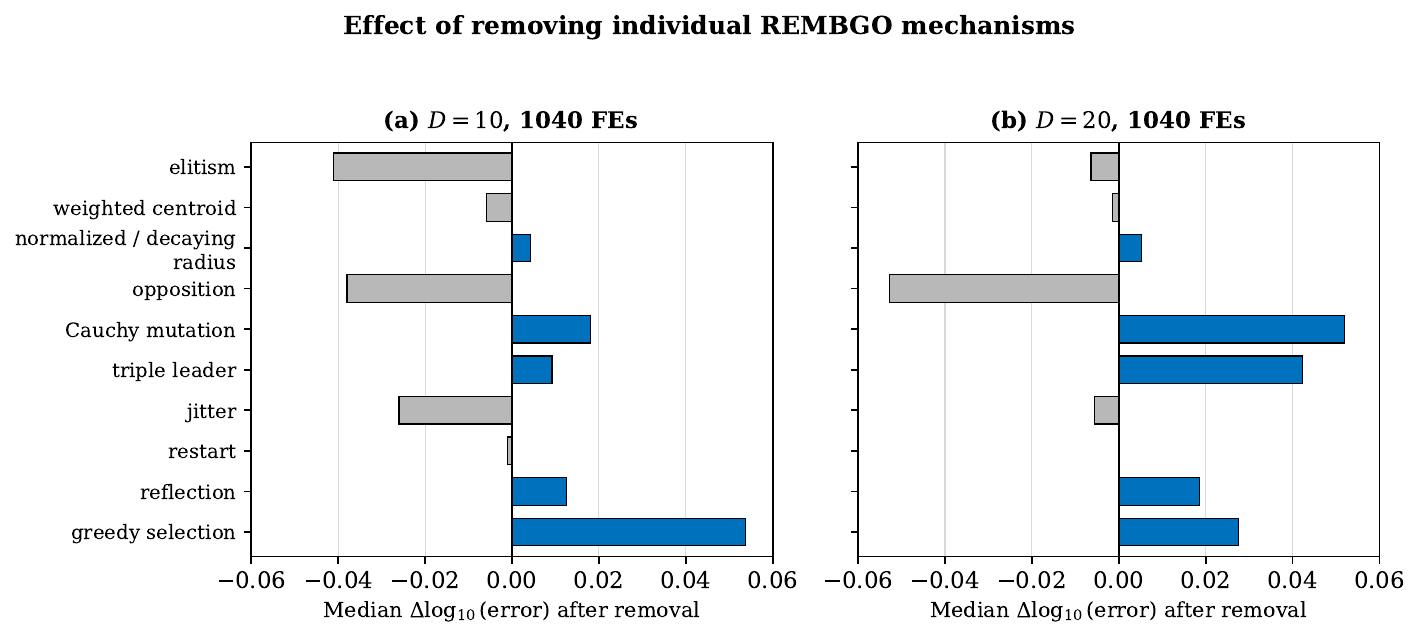}
\caption{Component-removal penalties for REMBGO at the exact $1040$-evaluation
budget ($D{=}10$ left, $D{=}20$ right). Positive values indicate degradation after
removal, negative a descriptive improvement; no removal is significant after the
suite-level Holm correction across the ten planned contrasts.}
\label{fig:mechanism}
\end{figure}

Table~\ref{tab:ablation_1040} reports an implementation-level mechanism-removal
study at the exact $1040$-evaluation budget, and Figure~\ref{fig:mechanism} shows
the
corresponding removal penalties; the removed elements include both retained and
refined parts of the complete REMBGO implementation. The evidence is
dimension-dependent. At $D{=}10$ the largest descriptive penalty follows removal
of greedy survivor selection (median $\Delta\log_{10}$ error $+0.054$, rank
$9.250$, Holm $p=0.093$), with removal of Cauchy mutation ($+0.018$) and of the
triple-leader move ($+0.009$) much smaller. At $D{=}20$ the most visible
descriptive penalties come instead from removing Cauchy mutation ($+0.052$, rank
$8.167$) and the triple-leader move ($+0.042$, rank $8.000$, Holm $p=0.068$), with
greedy selection ($+0.028$) still contributing. No individual removal is
significant after the suite-level Holm correction across the ten planned removal
contrasts at either dimension, and the pattern is more consistent with an
interacting search design than with a single dominant component.

A supplementary one-factor-at-a-time sensitivity analysis at the same
$1040$-evaluation budget uses six representative CEC2022 functions with twenty
matched runs per function at both dimensions (\SM{S2}), varying the Cauchy and
triple-leader probabilities, the elite count and the restart fraction. Nearby
settings can be descriptively better or worse than the defaults, so the selected
values are prespecified operating settings within a reasonably usable
neighbourhood, not claimed local optima, and the results do not justify post-hoc
retuning.

\begin{table}[tb]\centering
\caption{Component-removal results at the exact $1040$-evaluation budget; the
removed elements include both retained and refined parts of REMBGO. Positive median
$\Delta\log_{10}$ error indicates degradation after removal, negative a descriptive
improvement. No removal is significant after the suite-level Holm correction across
the ten planned contrasts.}
\label{tab:ablation_1040}
\small
\setlength\tabcolsep{6pt}\renewcommand{\arraystretch}{1.12}
\setlength\tabcolsep{3.4pt}
\begin{tabular}{lrrrrrr}
\toprule
Removed component & Rank $D{=}10$ & Rank $D{=}20$ & $\Delta\log_{10}$ $D{=}10$ &
$\Delta\log_{10}$ $D{=}20$ & Holm $p$ $D{=}10$ & Holm $p$ $D{=}20$ \\
\midrule
Elitism            & 3.583 & 4.833 & $-0.0411$ & $-0.0064$ & 0.617 & 1.000 \\
Weighted centroid  & 5.167 & 5.083 & $-0.0058$ & $-0.0015$ & 1.000 & 1.000 \\
Normalized/decaying radius & 6.667 & 6.500 & $+0.0043$ & $+0.0053$ & 1.000 & 1.000 \\
Opposition         & 3.917 & 2.583 & $-0.0380$ & $-0.0527$ & 0.617 & 0.098 \\
Cauchy mutation    & 7.583 & 8.167 & $+0.0181$ & $+0.0520$ & 1.000 & 1.000 \\
Triple leader      & 7.833 & 8.000 & $+0.0092$ & $+0.0423$ & 1.000 & 0.068 \\
Jitter             & 4.500 & 4.917 & $-0.0260$ & $-0.0056$ & 0.576 & 1.000 \\
Restart            & 4.833 & 5.167 & $-0.0011$ & $+0.0000$ & 1.000 & 1.000 \\
Reflection         & 7.083 & 6.833 & $+0.0125$ & $+0.0186$ & 1.000 & 1.000 \\
Greedy selection   & 9.250 & 8.417 & $+0.0537$ & $+0.0275$ & 0.093 & 0.083 \\
\bottomrule
\end{tabular}
\end{table}

\subsection{Fixed-gain EAF--STATCOM generalization}
The transfer study replays selected controllers on unseen furnace trajectories;
it is not an optimizer-ranking experiment. For each of the five optimizers and
each of the two controllers the gain vector with the lowest recorded
$J_\mathrm{tot}$ is selected (ten fixed vectors), mapped to controller parameters
by the same transformation used during tuning. Before the corrected campaign, a
separate provenance qualification replay reproduced the ten archived training
scores to a maximum absolute difference of $2.03\times10^{-8}$, confirming the
frozen gains and the objective implementation; that provenance check is not part
of the corrected transfer set. The corrected campaign then evaluates each of the
ten vectors on twenty pre-specified unseen arc realizations plus one corrected
base-$42$ reference realization, giving $210$ one-second EMT simulations in
total: $200$ unseen and $10$ reference. Every vector receives the identical twenty
unseen arcs and no optimizer is re-run. All $210$ cases are hard-feasible and
pass the delivered-configuration audit.

The arc realizations used here are generated by a corrected implementation of the
arc-length perturbation. Two stochastic conditions are distinguished here: STO-SR, the fixed-seed stochastic-mode realization that defined the original application tuning condition, and STO-ENS, the independent multi-realization stochastic ensemble used for frozen-gain transfer. STO-ENS is a methodological role and its realization count is paper-specific; here it is the twenty pre-specified unseen arc realizations, and the corrected base-$42$ realization is a separate reference case that is excluded from all STO-ENS statistics. Archived identifiers containing \texttt{STO} are retained in the source files for traceability. A post-campaign audit established that the sampler used during the original
campaign produced a periodic rather than broadband trajectory. The twenty unseen bases were therefore re-drawn under an
exact integer implementation and pre-specified before any corrected result was
computed; one candidate base was excluded by the pre-specified screen because one
of its phase seeds fell on the degenerate short orbit. No gain vector was retuned
and no vector was reselected. The corrected base-$42$ realization is reported as a
reference case only: because the generator itself changed, it is \emph{not} a
reproduction of the STO-SR tuning trajectory, and it is excluded from every
unseen-arc statistic below. The STO-SR training objective recorded during tuning
remains the reference against which train-to-unseen shifts are measured.

\begin{figure}[tbp]\centering
\includegraphics[width=\linewidth]{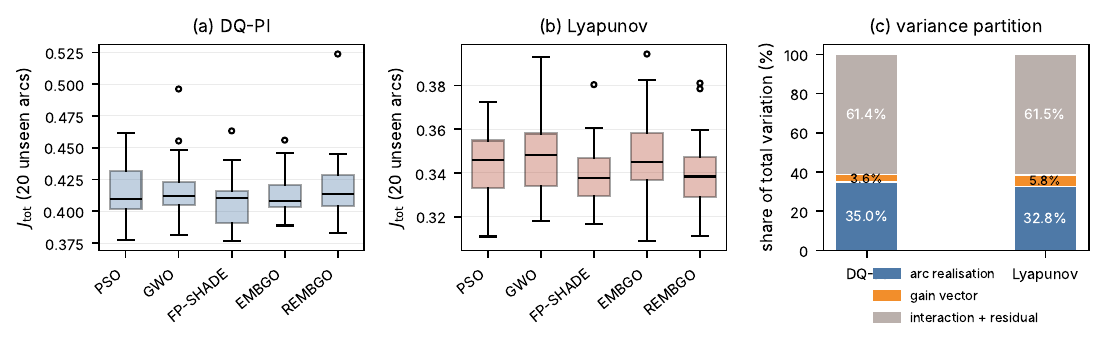}
\caption{Fixed-gain generalization across twenty unseen stochastic EAF arc
realizations for the five selected gain vectors of each controller. The first two
panels show the validation objective distributions for the DQ-PI and Lyapunov
controllers over the twenty pre-specified corrected arc realizations; the final
panel partitions the crossed variation into arc-realization, gain-vector and
interaction-plus-residual components. The corrected base-$42$ reference case is
excluded. No re-optimization is performed and no optimizer ranking is implied.}
\label{fig:eaf}
\end{figure}

The crossed-variation decomposition is more informative than the small
differences among the five validation means (Figure~\ref{fig:eaf}). Partitioning
the crossed arc-by-vector observations, the selected-vector main effect accounts
for only $3.64\%$ (DQ-PI) and $5.75\%$ (Lyapunov) of the total variation, the
arc-realization main effect for $34.98\%$ and $32.79\%$, and the arc-by-vector
interaction and residual for the remaining $61.38\%$ and $61.46\%$; arc realization
and especially the interaction dominate. Consistently, the five unseen-arc
validation means are close---for DQ-PI over $0.4071$--$0.4195$ and for Lyapunov
over $0.3384$--$0.3495$---differences that do not support declaring an optimizer
winner.

Averaged over the ten selected vectors, the shift from the STO-SR training
objective to the twenty-unseen-arc mean is $+0.0198$, with individual shifts from
$-0.0067$ to $+0.0419$. The shift is strongly controller-dependent: the five
DQ-PI vectors shift by only $+0.0059$ on average, and the FP-SHADE vector
improves slightly ($-0.0067$), whereas all five Lyapunov vectors degrade, by
$+0.0338$ on average. The Lyapunov vectors retain substantially lower absolute
$J_\mathrm{tot}$, so the larger relative degradation does not reverse their
objective ordering; it does show that lower tuned objective value and smaller
train-to-unseen shift are distinct properties.

The parent contrast is inconclusive on unseen arcs, and its direction depends on
the controller. The matched REMBGO$-$EMBGO difference (positive meaning the
REMBGO vector is worse) has, for DQ-PI, median $+0.001871$ and mean $+0.005956$
(signed-rank $p=0.498$, signs $8/0/12$, bootstrap $95\%$ CI
$[-0.0052,+0.0070]$, paired $t$ $p=0.388$) and, for Lyapunov, median $-0.006007$
and mean $-0.008843$ ($p=0.165$, signs $12/0/8$, CI $[-0.0196,+0.0044]$, paired
$t$ $p=0.104$); both intervals include zero. Neither controller family gives
evidence that the REMBGO-selected vector generalizes more or less consistently
than the EMBGO-selected vector across unseen trajectories. REMBGO is therefore
competitive with its parent under a strict fixed-gain unseen-arc transfer test,
with controller-dependent differences that are not statistically significant;
the algorithmic contribution rests on the benchmark and evaluation-budget
evidence of Section~\ref{sec:results} and \SM{S1}, not on an
application-side win.

% =============================================================================
\section{Discussion}\label{sec:disc}
Optimizer ordering varies materially with the evaluation budget---swarm methods
(GWO, PSO) lead at the smallest budgets, REMBGO is competitive at the exact
application budget, and FP-SHADE leads once many evaluations are available---so a
ranking established at a large benchmark budget need not describe performance near
$10^{3}$ evaluations. The parent comparison is the most direct evidence for the
modifications: REMBGO improves on EMBGO at the application budget ($4/6/2$ at
$D{=}10$, $9/1/2$ at $D{=}20$), with the strongest function-level parent advantage
at the smallest budgets and a reversal as the budget grows, so the refinements help
mainly in the evaluation-limited regime.
No removed component is significant after the suite-level Holm correction, so the
design reads as an interacting implementation rather than independently established
mechanisms. Beyond the Holm-corrected win/tie/loss counts, the archived 1040-evaluation
bridge records provide a descriptive magnitude measure for the parent contrast.
Across the twelve CEC2022 functions the median matched-pairs rank-biserial
correlation favouring REMBGO over EMBGO is $+0.35$ at $D=10$ and $+0.65$ at
$D=20$. The corresponding median per-function difference in median
$\log_{10}$ terminal error is $-0.06$ and $-0.11$ decades respectively, a
negative value indicating lower REMBGO error, and REMBGO has the lower median
terminal error on nine of the twelve functions at each dimension. These are
suite-level descriptive effect magnitudes and do not replace the Holm-corrected
function-wise inference.

The frozen REMBGO implementation retains an absolute centroid-eligibility test
$f_i<10^{7}$. This constant is inherited from an earlier application-oriented
implementation; it does not delimit the admissible region of the engineering
objective used here, whose nominal hard-penalty levels are of order $10^{4}$ and
$10^{5}$ and whose observed campaign values lie well below $10^{7}$. On raw
CEC2022 functions the objective is the benchmark value itself and its magnitude is
function dependent, so the same constant can become operationally active. A
side-effect-free audit at the primary budget reproduced all $720$ archived
terminal values exactly and found activation confined to F1, F6 and F8. A matched
sensitivity run that removed only the finite cutoff left every originally inactive
run bit-identical and produced no Holm-significant terminal-performance difference
in any affected function--dimension cell. The threshold is therefore
operationally active and objective-scale dependent on a subset of CEC2022
functions, but this analysis provides no evidence of a material
terminal-performance change from removing the finite cutoff at the tested
1040-evaluation condition, so the reported benchmark conclusions do not rest on
it. Supplementary Section~S6 gives the activation counts, the sensitivity design
and the per-cell results.

Recent analysis of the CEC2022 winner shows that added mechanisms and hidden parameters should not automatically be read as useful complexity, and that a simplified variant of that algorithm can outperform the original~\cite{ref_bied_simplify}. Component value should therefore be established empirically rather than asserted, which is why REMBGO is assessed here through a matched parent comparison, a component-removal study and a budget-dependent characterization rather than through terminal-rank claims. The component-removal result, in which no single removal survives multiplicity correction, is reported in that spirit: it indicates that the behaviour arises from interacting components and cannot be attributed confidently to any one mechanism. In the transfer test the
selected-vector main effect accounts for only $3.6\%$ (DQ-PI) and $5.8\%$
(Lyapunov) of the descriptive crossed validation variation, against $35.0\%$ and
$32.8\%$ for arc realization and approximately $61.4\%$ for interaction plus
residual, so
robustness across stochastic trajectories matters more than the small differences
among nominal tuning winners; this does not conflict with the benchmark parent
result, as the two experiments answer different questions.

% =============================================================================
\section{Limitations}\label{sec:limits}
The benchmark study covers the
CEC2022 suite only, at $D{=}10$ and $D{=}20$ with thirty runs per condition, and
Holm correction is applied within each budget without cross-budget multiplicity
control, so the win/tie/loss counts describe breadth and direction rather than
magnitude. The application study uses one EAF--STATCOM plant family, two
controllers, a single tuning trajectory and twenty unseen stochastic arcs,
evaluated by simulation replay rather than on hardware. Finally, two implementation
details---range-normalization of the neighbourhood radius and the midpoint-based
restart subgroup---have limited identifiability on the symmetric isotropic CEC2022
box, though matched diagnostics found no detectable change in the reported
conclusions.

% =============================================================================

Benchmark breadth is a further limitation. CEC2022 contains twelve functions per
dimension, which is small by the standards of recent benchmarking
methodology~\cite{ref_piotrowski3}: although the present design compensates
partly through thirty repetitions, two dimensions, several evaluation budgets, a
matched parent comparison, a component-removal study and an independent
engineering transfer test, cross-suite validation remains necessary before
broader optimizer-ranking claims are made. Published analyses also show that the
ranking method and the tuning effort applied to competing algorithms can
themselves change which optimizer appears best on this suite~\cite{ref_bied3},
which is one reason the present evidence is reported as a matched parent
comparison and a budget characterization rather than as a terminal rank.

\section{Conclusion}\label{sec:concl}
Optimizer ordering depends materially on the evaluation budget, so optimizer
selection for simulation-expensive controller tuning should use a budget
representative of the intended application. REMBGO is not universally best: at $1040$
evaluations it holds the lowest observed mean rank at $D{=}10$ and GWO at $D{=}20$,
while FP-SHADE leads at large budgets. Selecting a nominal winner on one
stochastic trajectory does not by itself establish robust generalization. In the
corrected transfer study all ten frozen champions remained feasible across twenty
pre-specified unseen stochastic arcs. The average shift from the STO-SR training
objective to the unseen-arc mean was $+0.0198$, with smaller mean degradation for
the DQ-PI vectors ($+0.0059$) than for the Lyapunov vectors ($+0.0338$), and the
arc realization and its interaction with the selected vector accounted for far
more of the validation variation than the choice of vector itself. The matched
REMBGO$-$EMBGO paired differences were controller-dependent in sign and not
statistically significant for either architecture, so the EAF experiment supports
fixed-gain transfer robustness rather than an application-side REMBGO
superiority claim. Future work should examine multi-trajectory or risk-aware
controller tuning under the same restricted evaluation-budget setting.

% =============================================================================

\paragraph{Data availability.}
The numerical results and methodological details supporting the reported findings are provided in the article and supplementary material.

\end{document}

\setcounter{section}{0}\setcounter{table}{0}\setcounter{figure}{0}\setcounter{equation}{0}
\renewcommand{\thesection}{S\arabic{section}}
\renewcommand{\thetable}{S\arabic{table}}
\renewcommand{\thefigure}{S\arabic{figure}}
\renewcommand{\theequation}{S\arabic{equation}}
\begin{center}{\large\bfseries Supplementary Material}\end{center}
\medskip

\noindent This supplement collects the secondary material referenced from the
main paper. Section~S1 gives the complete evaluation-budget ladder, Section~S2
the sensitivity study at the exact $1040$-evaluation budget, Section~S3
additional component diagnostics, Section~S4 the full fixed-gain
EAF--STATCOM generalization results, and Section~S5 gives the reproducibility and
statistical conventions together with a compact application-objective summary.
Every table and figure regenerates
from the archived CEC2022 and application result files without rerunning any
optimizer.

% =============================================================================
\section{Complete five-optimizer evaluation-budget ladder}\label{sm:ladder}
Table~\ref{tab:sm_ladder} lists the full twenty-eight-rung ladder underlying
Figure~1 and Table~2 of the main paper: the Friedman mean rank of each optimizer
(lower is better), the optimizer with the lowest observed mean rank, the per-rung
omnibus Friedman $p$-value, and the REMBGO-versus-EMBGO function-level record
(REMBGO better / no significant difference / EMBGO better) after Holm correction
across the twelve CEC2022 functions at that rung. The $1040$-evaluation rung is the
exact application-aligned budget. The Holm family is the twelve functions within a
single rung; cross-rung multiplicity is not controlled, so the rungs are not
compared inferentially with one another.

\begin{table}[H]\centering
\caption{Complete five-optimizer CEC2022 evaluation-budget ladder. Mean Friedman ranks,
lowest-mean-rank optimizer, per-rung omnibus Friedman $p$, and Holm-corrected
REMBGO-vs-EMBGO win/tie/loss over the twelve functions. Source: archived CEC2022
budget-ladder result set.}
\label{tab:sm_ladder}
\setlength\tabcolsep{4.5pt}\renewcommand{\arraystretch}{1.08}\footnotesize
\begin{tabular}{rrrrrrrlcc}
\toprule
$D$ & FEs & PSO & GWO & FP-SHADE & EMBGO & REMBGO & Best & Friedman $p$ & W/T/L\\
\midrule
10 & 200    & 1.917 & 1.917 & 4.083 & 4.417 & 2.667 & PSO/GWO  & 1.93e-05 & 10/2/0\\
10 & 316    & 2.417 & 1.750 & 4.167 & 4.083 & 2.583 & GWO      & 1.89e-04 & 8/4/0\\
10 & 502    & 2.583 & 2.083 & 4.000 & 3.917 & 2.417 & GWO      & 4.06e-03 & 6/5/1\\
10 & 796    & 2.417 & 2.333 & 4.000 & 3.500 & 2.750 & GWO      & 0.0393   & 3/7/2\\
10 & 840    & 2.583 & 2.500 & 4.083 & 3.417 & 2.417 & REMBGO   & 0.0382   & 3/7/2\\
10 & 1040   & 2.750 & 2.833 & 4.000 & 3.167 & 2.250 & REMBGO   & 0.0892   & 4/6/2\\
10 & 1261   & 2.667 & 2.833 & 3.833 & 3.000 & 2.667 & PSO      & 0.339    & 3/8/1\\
10 & 2000   & 3.083 & 3.083 & 3.583 & 2.833 & 2.417 & REMBGO   & 0.483    & 2/7/3\\
10 & 3169   & 3.000 & 3.750 & 2.083 & 2.833 & 3.333 & FP-SHADE & 0.116    & 1/8/3\\
10 & 5023   & 2.750 & 4.583 & 1.667 & 2.833 & 3.167 & FP-SHADE & 2.98e-04 & 0/8/4\\
10 & 7962   & 3.000 & 4.833 & 1.333 & 2.417 & 3.417 & FP-SHADE & 1.97e-06 & 1/6/5\\
10 & 12619  & 2.917 & 4.917 & 1.167 & 2.583 & 3.417 & FP-SHADE & 3.72e-07 & 0/6/6\\
10 & 20000  & 2.833 & 4.917 & 1.292 & 2.458 & 3.500 & FP-SHADE & 5.06e-07 & 0/7/5\\
10 & 200000 & 2.542 & 4.917 & 1.625 & 2.625 & 3.292 & FP-SHADE & 1.07e-06 & 0/9/3\\
\midrule
20 & 125    & 2.250 & 1.250 & 4.583 & 4.333 & 2.583 & GWO      & 7.66e-08 & 10/2/0\\
20 & 227    & 2.083 & 1.417 & 4.417 & 4.417 & 2.667 & GWO      & 3.08e-07 & 11/1/0\\
20 & 414    & 2.250 & 1.333 & 4.167 & 4.583 & 2.667 & GWO      & 4.36e-07 & 9/3/0\\
20 & 754    & 2.250 & 1.750 & 3.750 & 4.167 & 3.083 & GWO      & 6.36e-04 & 7/4/1\\
20 & 840    & 2.417 & 1.583 & 3.917 & 4.167 & 2.917 & GWO      & 2.13e-04 & 8/3/1\\
20 & 1040   & 2.500 & 2.250 & 3.833 & 3.833 & 2.583 & GWO      & 0.0224   & 9/1/2\\
20 & 1373   & 3.000 & 2.250 & 3.500 & 3.333 & 2.917 & GWO      & 0.347    & 7/3/2\\
20 & 2500   & 3.250 & 2.750 & 2.417 & 3.417 & 3.167 & FP-SHADE & 0.525    & 5/5/2\\
20 & 4551   & 3.083 & 3.750 & 1.500 & 3.000 & 3.667 & FP-SHADE & 3.50e-03 & 0/11/1\\
20 & 8286   & 3.000 & 4.250 & 1.167 & 2.667 & 3.917 & FP-SHADE & 1.14e-05 & 0/9/3\\
20 & 15085  & 2.667 & 4.667 & 1.083 & 2.750 & 3.833 & FP-SHADE & 4.36e-07 & 0/8/4\\
20 & 27464  & 2.583 & 4.667 & 1.000 & 2.583 & 4.167 & FP-SHADE & 3.05e-08 & 0/4/8\\
20 & 50000  & 3.000 & 4.917 & 1.000 & 2.417 & 3.667 & FP-SHADE & 3.25e-08 & 0/5/7\\
20 & 1000000& 2.542 & 4.917 & 1.542 & 2.875 & 3.125 & FP-SHADE & 3.82e-06 & 0/8/4\\
\bottomrule
\end{tabular}
\end{table}

% =============================================================================
\section{Sensitivity at the exact $1040$-evaluation budget}\label{sm:sens}
\begin{figure}[H]\centering
\includegraphics[width=0.9\textwidth]{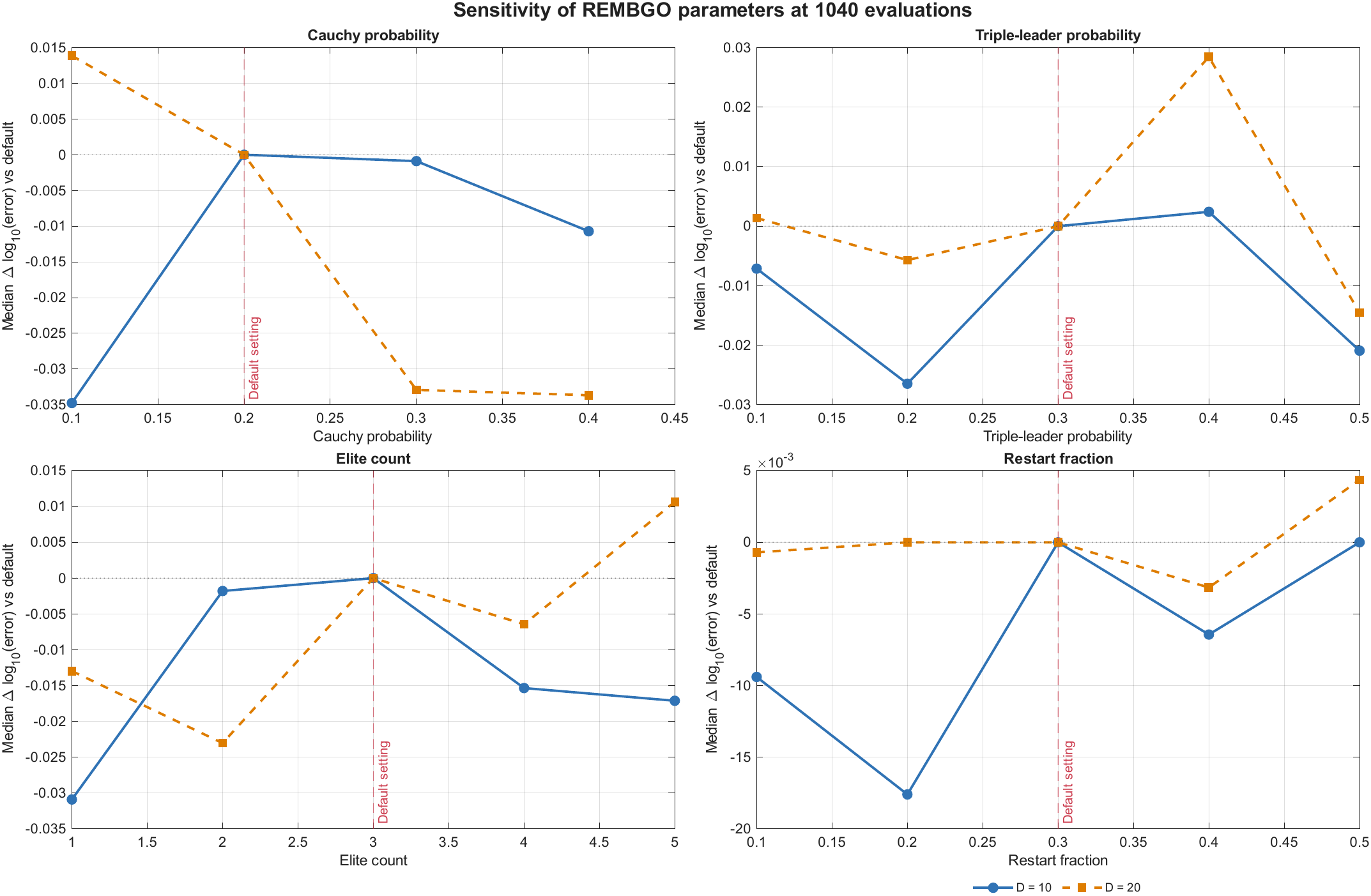}
\caption{REMBGO one-factor-at-a-time sensitivity at the exact $1040$-evaluation
budget for $D{=}10$ and $D{=}20$. The selected default of each factor is marked;
nearby settings are descriptively better or worse without a consistent
improvement across both dimensions.}
\label{fig:sm_sens}
\end{figure}

A one-factor-at-a-time sweep around the prespecified defaults was run at the exact
$1040$-evaluation budget for both dimensions, varying the Cauchy probability, the
triple-leader probability, the elite count and the restart fraction while holding
all other settings at Table~1 values. The sweep uses six representative CEC2022
functions (F1, F4, F6, F7, F9, F12) with twenty matched runs each. This subset is
chosen to span the four CEC2022 landscape classes at a fraction of the full-suite
cost---F1 (unimodal), F4 (basic multimodal), F6 and F7 (hybrid), and F9 and F12
(composition)---so that the sensitivity sweep probes each landscape type while
keeping the four-factor grid tractable; it is deliberately smaller than the
thirty-run, twelve-function protocol used for the confirmatory benchmark and
component-removal studies, and is reported as an exploratory robustness check
rather than a tuning procedure. For each
dimension and swept factor, the settings of that factor are ranked separately on
every one of the six functions using the median terminal error over the twenty
runs, and the reported rank is the mean of these ranks across the six functions;
the displayed median $\Delta\log_{10}$ is the median over the six functions of the
change in $\log_{10}$ terminal error relative to the default setting of that
factor. Table~\ref{tab:sm_sens} gives the mean rank and this median
$\Delta\log_{10}$, and Figure~\ref{fig:sm_sens} shows the same sweep. Nearby values can be
descriptively better or worse than the selected defaults, and no configuration
dominates across both dimensions. The defaults therefore sit within a reasonably
usable neighbourhood and are retained as prespecified operating settings rather
than retuned post hoc.

\begin{table}[H]\centering
\caption{REMBGO sensitivity at the exact $1040$-evaluation budget. Within each
swept factor the ranks are computed across that factor's settings; the default
value (marked $^\ast$) defines the zero of $\Delta\log_{10}$. Source: archived
$1040$-evaluation sensitivity result set.}
\label{tab:sm_sens}
\setlength\tabcolsep{6pt}\renewcommand{\arraystretch}{1.1}\footnotesize
\begin{tabular}{llcccc}
\toprule
\textbf{Parameter} & \textbf{Value} & \textbf{Rank $D{=}10$} &
\textbf{$\Delta\log_{10}$ $D{=}10$} & \textbf{Rank $D{=}20$} &
\textbf{$\Delta\log_{10}$ $D{=}20$}\\ \midrule
\multirow{4}{*}{Cauchy probability}
 & 0.1        & 2.167 & $-0.0348$ & 2.833 & $+0.0139$\\
 & $0.2^\ast$ & 2.833 & $\phantom{+}0.0000$ & 3.000 & $\phantom{+}0.0000$\\
 & 0.3        & 3.000 & $-0.0009$ & 2.167 & $-0.0329$\\
 & 0.4        & 2.000 & $-0.0107$ & 2.000 & $-0.0337$\\ \midrule
\multirow{5}{*}{Elite count}
 & 1        & 2.500 & $-0.0309$ & 2.500 & $-0.0130$\\
 & 2        & 2.667 & $-0.0018$ & 1.667 & $-0.0230$\\
 & $3^\ast$ & 3.833 & $\phantom{+}0.0000$ & 3.333 & $\phantom{+}0.0000$\\
 & 4        & 2.500 & $-0.0153$ & 2.833 & $-0.0064$\\
 & 5        & 3.500 & $-0.0171$ & 4.667 & $+0.0106$\\ \midrule
\multirow{5}{*}{Restart fraction}
 & 0.1        & 2.250 & $-0.0094$ & 3.000 & $-0.0007$\\
 & 0.2        & 2.250 & $-0.0176$ & 2.750 & $\phantom{+}0.0000$\\
 & $0.3^\ast$ & 3.750 & $\phantom{+}0.0000$ & 3.083 & $\phantom{+}0.0000$\\
 & 0.4        & 3.000 & $-0.0064$ & 2.917 & $-0.0032$\\
 & 0.5        & 3.750 & $\phantom{+}0.0000$ & 3.250 & $+0.0044$\\ \midrule
\multirow{5}{*}{Triple-leader probability}
 & 0.1        & 2.833 & $-0.0071$ & 3.167 & $+0.0014$\\
 & 0.2        & 2.667 & $-0.0265$ & 3.000 & $-0.0057$\\
 & $0.3^\ast$ & 3.500 & $\phantom{+}0.0000$ & 3.000 & $\phantom{+}0.0000$\\
 & 0.4        & 3.667 & $+0.0024$ & 4.167 & $+0.0284$\\
 & 0.5        & 2.333 & $-0.0209$ & 1.667 & $-0.0145$\\
\bottomrule
\end{tabular}
\end{table}

% =============================================================================
\section{Additional REMBGO component diagnostics}\label{sm:mech}
\paragraph{Restart-reference behaviour on symmetric bounds.}
The partial restart (R11) regenerates part of its group as multiplicative
neighbours of a reference point $\bm{x}_\mathrm{ref}$. On the symmetric CEC2022
bounds $\bm{x}_\mathrm{ref}$ defaults to the box midpoint, which is the origin, so
the multiplicative rule $\bm{x}_\mathrm{ref}\odot(1+0.3\,\mathcal U[-1,1]^D)$
returns the origin unchanged and that subgroup injects identical zero vectors. A
matched diagnostic replaced the reference subgroup with an alternative
non-degenerate reference and re-ran REMBGO under identical seeds; the alternative
reference produced no detectable change on the isotropic benchmark. Failure to
detect a difference is not evidence of equality; the reported benchmark results are
stated as conditional on the implemented behaviour, and this diagnostic does not
alter the parent-relative records or the application study, which never uses the
symmetric-box reference.

\paragraph{Coordinate-normalization cancellation.}
The normalized decaying neighbourhood radius (R3) combines range normalization
with the temporal decay $\rho(\tau)=1.5-0.7\tau$. For the idealized rule with
$\varepsilon_R=0$, if every coordinate has the same range $s$ then
$R_0=\lVert\bm x_b-\bm x_w\rVert_2\,\delta\,\rho(\tau)/s$ and the membership
condition $\lVert\bm x_i-\bm x_b\rVert_2/s\le R_0$ reduces exactly to
$\lVert\bm x_i-\bm x_b\rVert_2\le\lVert\bm x_b-\bm x_w\rVert_2\,\delta\,\rho(\tau)$,
so the scale cancels. The implemented form keeps the regularizer
$\varepsilon_R=10^{-12}$ inside the norm; nevertheless, on the isotropic CEC2022
box the runs with and without range normalization were bit-identical for every
tested seed, function and budget. Thus, only the temporal-decay contribution is
active in these isotropic CEC2022 tests, and range normalization neither helps nor
harms; it is retained because it becomes active on anisotropic domains outside the
CEC2022 specification.

% =============================================================================
\section{Fixed-gain EAF--STATCOM generalization: full results}\label{sm:arc}
Table~\ref{tab:sm_gains} gives the ten selected minimum-$J_\mathrm{tot}$ gain
vectors that were replayed, so that the fixed controllers used in the transfer
study are fully specified. Table~\ref{tab:sm_arcgen} summarises the per-vector
validation performance of these ten vectors across the twenty unseen arcs,
Table~\ref{tab:sm_paired} gives the matched REMBGO$-$EMBGO paired statistics, and
Table~\ref{tab:sm_decomp} the crossed-variation decomposition. Each row of
Table~\ref{tab:sm_arcgen} is one fixed selected vector replayed on twenty matched
trajectories, not an optimizer sample; the paired and decomposition results are
descriptive of these post-selected champions and are not optimizer comparisons.

\begin{table}[H]\centering
\caption{Selected minimum-$J_\mathrm{tot}$ gain vectors replayed in the
fixed-gain transfer study, one per optimizer and controller. For DQ-PI the
reactive integral gain follows $K_{i,Q}=K_{p,Q}/T_{i,Q}$ and is not listed
separately. Values are the archived stochastic-tuning champions; each reproduces
the tabulated $J_\mathrm{train}$ of Table~\ref{tab:sm_arcgen} on replay.}
\label{tab:sm_gains}
\setlength\tabcolsep{5pt}\renewcommand{\arraystretch}{1.1}\footnotesize
\begin{subtable}{\linewidth}\centering
\caption{DQ-PI controller: $[K_{p,id},K_{i,id},K_{p,iq},K_{i,iq},K_{p,Q},T_{i,Q}]$.}
\begin{tabular}{lrrrrrr}
\toprule
\textbf{Optimizer} & $K_{p,id}$ & $K_{i,id}$ & $K_{p,iq}$ & $K_{i,iq}$ &
$K_{p,Q}$ & $T_{i,Q}$\\ \midrule
PSO      & 0.3078 & 184.01 & 1.3728 & 610.54 & $-27.988$ & \phantom{0}8.0852\\
GWO      & 0.3252 & 176.00 & 1.4500 & 199.13 & $-25.595$ & \phantom{0}6.0716\\
FP-SHADE & 0.3282 & 211.96 & 1.3744 & 539.70 & $-27.274$ & 11.6891\\
EMBGO    & 0.3368 & 210.96 & 1.3565 & 558.23 & $-27.176$ & 16.8587\\
REMBGO   & 0.3392 & 155.22 & 1.4253 & 502.85 & $-32.227$ & \phantom{0}5.9562\\
\bottomrule
\end{tabular}
\end{subtable}

\vspace{4pt}
\begin{subtable}{\linewidth}\centering
\caption{Lyapunov controller: $[K_\mathrm{droop},K_{q,p},K_{q,i},k_{1,i},k_{i,i},k_{2,i}]$.}
\begin{tabular}{lrrrrrr}
\toprule
\textbf{Optimizer} & $K_\mathrm{droop}$ & $K_{q,p}$ & $K_{q,i}$ & $k_{1,i}$ &
$k_{i,i}$ & $k_{2,i}$\\ \midrule
PSO      & 29489.96 & 5.8962 & 2872.73 & 3068.29 & 67.152 & 1264.43\\
GWO      & 33574.82 & 5.3477 & 3000.00 & 4084.43 & 19.121 & \phantom{0}523.58\\
FP-SHADE & 22136.93 & 8.0899 & 2972.88 & 2708.44 & 11.810 & 2165.53\\
EMBGO    & 25459.80 & 7.5006 & 2973.74 & 3384.57 & \phantom{0}5.000 & 1277.77\\
REMBGO   & 24540.15 & 8.2296 & 2993.97 & 3319.55 & 48.010 & 1250.57\\
\bottomrule
\end{tabular}
\end{subtable}
\end{table}

\begin{table}[H]\centering
\caption{Fixed-gain replay of the ten selected minimum-$J_\mathrm{tot}$ vectors.
$J_\mathrm{train}$ is the tuning-arc score; validation columns summarise
$J_\mathrm{tot}$ over the twenty pre-registered corrected unseen arcs; mean shift
is validation mean minus $J_\mathrm{train}$, where $J_\mathrm{train}$ is the
STO-SR tuning-arc score recorded during the original campaign. The final column
is the corrected base-$42$ realization, reported as a reference case only and
excluded from every validation statistic; it is not a reproduction of the STO-SR
tuning trajectory. Source: the corrected-generator campaign dataset, 200 unseen rows.}
\label{tab:sm_arcgen}
\setlength\tabcolsep{3.6pt}\renewcommand{\arraystretch}{1.1}\scriptsize
\begin{tabular}{ll cccccc}
\toprule
\textbf{Controller} & \textbf{Optimizer} & $J_\mathrm{train}$ & Val.\ mean
& Val.\ SD & Val.\ median & Mean shift & Corr.\ base-42\\ \midrule
\multirow{5}{*}{DQ-PI}
 & PSO      & 0.406577 & 0.416711 & 0.023752 & 0.409602 & $+0.010133$ & 0.402806\\
 & GWO      & 0.402520 & 0.419252 & 0.025181 & 0.411996 & $+0.016732$ & 0.419992\\
 & FP-SHADE & 0.413838 & 0.407128 & 0.021850 & 0.410386 & $-0.006710$ & 0.388682\\
 & EMBGO    & 0.412499 & 0.413542 & 0.017652 & 0.408077 & $+0.001044$ & 0.426294\\
 & REMBGO   & 0.411379 & 0.419498 & 0.030625 & 0.413860 & $+0.008119$ & 0.399341\\ \midrule
\multirow{5}{*}{Lyapunov}
 & PSO      & 0.310776 & 0.343695 & 0.016917 & 0.345919 & $+0.032919$ & 0.353066\\
 & GWO      & 0.311053 & 0.349493 & 0.020933 & 0.348313 & $+0.038440$ & 0.358254\\
 & FP-SHADE & 0.312073 & 0.338408 & 0.015740 & 0.337521 & $+0.026335$ & 0.324189\\
 & EMBGO    & 0.306325 & 0.348196 & 0.020400 & 0.344901 & $+0.041871$ & 0.315469\\
 & REMBGO   & 0.309969 & 0.339353 & 0.018601 & 0.338392 & $+0.029383$ & 0.341848\\
\bottomrule
\end{tabular}
\end{table}

\begin{table}[H]\centering
\caption{Matched REMBGO$-$EMBGO fixed-gain paired difference across the twenty
unseen arcs (positive means the REMBGO vector has the larger, worse
$J_\mathrm{tot}$). CI is a $10{,}000$-resample percentile bootstrap; $p$ is the
two-sided exact signed-rank test; signs are REMBGO better / ties / REMBGO worse.
Source: the corrected-generator campaign dataset.}
\label{tab:sm_paired}
\setlength\tabcolsep{6pt}\renewcommand{\arraystretch}{1.1}\footnotesize
\begin{tabular}{lccccc}
\toprule
\textbf{Controller} & Mean & Median & $95\%$ CI & Signed-rank $p$ & Signs\\ \midrule
DQ-PI    & $+0.005956$ & $+0.001871$ & $[-0.005184,+0.007000]$ & 0.4980 & 8/0/12\\
Lyapunov & $-0.008843$ & $-0.006007$ & $[-0.019601,+0.004355]$ & 0.1650 & 12/0/8\\
\bottomrule
\end{tabular}
\end{table}

\begin{table}[H]\centering
\caption{Crossed-variation decomposition of the fixed-gain validation objective
over the twenty unseen arcs and five selected vectors. Percentages are shares of
the total sum of squares; the fixed-vector Friedman test and Kendall's $W$ are
descriptive checks over post-selected champions. Source: the corrected-generator
campaign dataset.}
\label{tab:sm_decomp}
\setlength\tabcolsep{6pt}\renewcommand{\arraystretch}{1.1}\footnotesize
\begin{tabular}{lccccc}
\toprule
\textbf{Controller} & Arc realization & Gain vector & Interaction $+$ residual &
Fixed-vector Friedman $p$ & Kendall $W$\\ \midrule
DQ-PI    & 34.98\% & 3.64\% & 61.38\% & 0.641 & 0.031\\
Lyapunov & 32.79\% & 5.75\% & 61.46\% & 0.585 & 0.036\\
\bottomrule
\end{tabular}
\end{table}

% =============================================================================
\section{Reproducibility, statistical conventions and compact objective summary}\label{sm:repro}
\paragraph{Seed and evaluation-accounting conventions.}
For each (dimension, function, run) the initial population is drawn
uniform-randomly over $[\bm l,\bm u]$ from the official CEC2022
\texttt{Rand\_Seeds} list with the prescribed Mersenne-Twister generator, and the
identical population is supplied to every optimizer, so all comparisons are
matched. Every objective call is counted identically across optimizers. A complete
$T$-generation run with population $N$ costs $N(T+1)$ evaluations including
initialization. For a requested budget $B$, $T_B=\lceil(B-N)/N\rceil$ sets the
maximum generation count, and the final generation is truncated when necessary so
that exactly $B$ objective calls are consumed. In a truncated final generation the
trial rows are evaluated in implementation order, and REMBGO does not reshuffle the
copied elite rows before truncation. Per-function differences are formed per seed
as
$d=f_{\mathrm{EMBGO}}-f_{\mathrm{REMBGO}}$; exact zero differences are dropped
following Wilcoxon's rule and the effective sample size reduced accordingly; tied
absolute differences receive average ranks; the test is the two-sided exact
signed-rank test; the direction is taken from the signed-rank statistic; and Holm
correction is applied across the twelve functions, one family per budget. Seeds
are never pooled across functions.

\paragraph{Component-removal statistical conventions.}
The component-removal study reported in the main paper is run at the exact
$1040$-evaluation budget on the same matched CEC2022 records, separately for
$D{=}10$ and $D{=}20$. Eleven configurations are compared: the full REMBGO
configuration and its ten implementation-level mechanism-removal variants. Each of the twelve CEC2022
functions is a statistical block; within a block the per-configuration observation
is the median terminal error over the thirty matched runs. The mean rank reported
for a removal is obtained by ranking the eleven configurations on each function by
this median error and averaging the rank across the twelve functions, so the full
REMBGO configuration also carries a rank and a removal that improves on it scores a
lower rank. The displayed median $\Delta\log_{10}$ is the median over the twelve
functions of $\log_{10}$(removal median error)$-\log_{10}$(full-REMBGO median
error), so a positive value means the removal degrades performance. Two Wilcoxon
signed-rank families are distinguished. An exploratory per-function family applies
the two-sided exact signed-rank test to the thirty matched pairs of each function,
with Holm correction across the twelve functions for each removal. The suite-level
family---the ``Holm $p$'' reported in Table~3 and the basis for the statement that
no removal is significant after the suite-level Holm correction---applies the exact
signed-rank test across the twelve per-function $\log_{10}$ medians for each
full-versus-removal contrast and then Holm-corrects across the ten planned removal
contrasts, separately for each dimension. Zero and tie handling follows the main
convention above, and each per-function median error is floored at $10^{-8}$ before
the base-10 logarithm.

These are implementation-level ablations. Because REMBGO uses sequential branch
selection under a fixed objective-call budget, disabling one branch redistributes
search exposure among the remaining branches, and removing elitism also changes the
number of newly generated candidates per generation. The contrasts therefore
quantify the effect of removing each component from the integrated implementation,
rather than an isolated causal contribution obtained with all other operator
exposures held fixed; this reinforces the reading of REMBGO as an interacting
search design.

\paragraph{L\'evy step scale.}
The L\'evy escape (R7) uses the Mantegna scale for exponent $\beta_L=1.5$,
\begin{equation}
\sigma_L=\left[\frac{\Gamma(1+\beta_L)\,\sin(\pi\beta_L/2)}
{\Gamma\!\bigl((1+\beta_L)/2\bigr)\,\beta_L\,2^{(\beta_L-1)/2}}\right]^{1/\beta_L},
\end{equation}
matching the implemented value, so that the main-text L\'evy update is fully
reproducible.

\paragraph{Compact application-objective summary.}
The scalar objective $J_\mathrm{tot}$ used in both the tuning application and the
fixed-gain transfer study is evaluated on the post-transient window
$t\in[0.50,1.00]$~s at the recorded sample rate, with fixed feature configuration:
fundamental $f_0=50$~Hz, envelope target rate $500$~Hz, voltage low-pass cutoff
$35$~Hz, voltage evaluation band $[7,12]$~Hz, reactive-power band $[5,15]$~Hz and
moving-standard-deviation window $0.10$~s. Two performance features are formed on
this window. The reactive-power fluctuation feature $Q_\mathrm{fluc}$ combines the
detrended standard deviation and the $[5,15]$~Hz band-pass RMS of the instantaneous
three-phase reactive power; the voltage feature $V_\mathrm{pcc}$ is the worst-phase
value of a per-cycle-RMS PCC-voltage composite built from the low-pass ($35$~Hz)
and band-pass ($[7,12]$~Hz) RMS of the relative voltage deviation plus a
moving-standard-deviation term. Each feature is normalized by its value on a common
uncompensated deterministic reference run, giving
$T_Q=Q_\mathrm{fluc}/Q_\mathrm{fluc}^\mathrm{ref}$ and
$T_V=V_\mathrm{pcc}/V_\mathrm{pcc}^\mathrm{ref}$, and the performance term is
$J_\mathrm{perf}=0.60\,T_Q+0.40\,T_V$.

Graded quality penalties act quadratically on the window-mean reactive-power draw
beyond a small allowance ($P_{\overline Q}$, weight $0.05$, referenced to a
$15$~kVAr allowance) and on the dc-link peak deviation beyond a $3\%$ band
($P_{V_\mathrm{dc}}$, weight $0.03$, $2\%$ scale). The safety term
$P_\mathrm{safety}=P_\mathrm{hard}+P_\mathrm{sat}+P_\mathrm{dq}$ combines hard
feasibility gates---triggered when the mean or peak reactive power exceeds the
$Q_\mathrm{base}=500$~kVAr compensator rating, when the dc-link peak deviation
exceeds $8\%$, or when the dc-link voltage leaves
$[0.80,1.30]\,V_\mathrm{dc}^\mathrm{ref}$, each adding a large fixed constant
($10^{4}$ for the reactive-power gates, $10^{5}$ for the dc-link gate) that renders
the candidate effectively infeasible---with graded soft costs on converter
over-modulation, $P_\mathrm{sat}=10\max(0,\mathrm{sat}-0.08)$, and current-tracking
error, $P_\mathrm{dq}=10\max(0,\mathrm{trk}-0.10)$. A candidate is hard-feasible
when no hard gate fires, so $P_\mathrm{hard}=0$. The assembled objective is
\begin{equation}
J_\mathrm{tot}=J_\mathrm{perf}+P_{\overline Q}+P_{V_\mathrm{dc}}
+P_\mathrm{hard}+P_\mathrm{sat}+P_\mathrm{dq},
\label{eq:sm_jtot}
\end{equation}
identical in both the application and the transfer study. All $210$ replayed
transfer cases are hard-feasible; the DQ-PI cases carry a nonzero
$P_{V_\mathrm{dc}}$ from a $5.675$--$5.979\%$ dc-link peak (above the $3\%$
free-quality band but below the $8\%$ hard gate), a characteristic of the DQ-PI
structure already present on the tuning arc, while every Lyapunov case meets the
stricter zero-penalty screen.

\section{Threshold audit and finite-cutoff sensitivity}\label{sms:fctr}

\paragraph{Origin of the constant.} The test is not a tuned search parameter and
does not mark the boundary of an admissible region; it is an absolute
centroid-eligibility threshold in the frozen implementation. Two facts are established by direct inspection. The same
$10^{7}$ constant appears as the terminal cap in the superseded v3 application
objectives, and it remains as the centroid-eligibility threshold in the frozen
REMBGO core. The authoritative objective chain used for the present work
contains no such cap and is not mathematically bounded by $10^{7}$; its nominal
hard-penalty constants are of order $10^{4}$ (reactive-power excursion) and
$10^{5}$ (dc-link excursion). The threshold therefore lies far above the
observed campaign objective values and those nominal constants, and was
effectively inactive on the application-scale runs examined here. CEC2022
objectives are raw benchmark values with function-dependent magnitudes, which
is why the same constant filters candidates on large-magnitude functions. The
constant is inherited from the application-oriented implementation rather than
being a scale-invariant feature of the search.

\paragraph{Instrumented audit.} A diagnostic clone of the frozen core was built
by inserting counters only; a static guard confirmed that the number of
random-number and objective call sites was unchanged, and all $720$ runs
reproduced their archived terminal values exactly with $1040$ evaluations and
$25$ iterations, which is the operative proof that the instrumentation is
passive. Activation by function, as a percentage of the $30\,000$
centroid-candidate observations in each cell:

\begin{center}\footnotesize
\begin{tabular}{lrrrrrrrrrrrr}
\hline
 & F1 & F2 & F3 & F4 & F5 & F6 & F7 & F8 & F9 & F10 & F11 & F12\\
\hline
$D{=}10$ & 19.8 & 0 & 0 & 0 & 0 & 63.0 & 0 & 3.6 & 0 & 0 & 0 & 0\\
$D{=}20$ & 38.5 & 0 & 0 & 0 & 0 & 96.1 & 0 & 4.4 & 0 & 0 & 0 & 0\\
\hline
\end{tabular}
\end{center}

\noindent Totals: $67\,661$ of $707\,724$ finite candidate observations
($9.56\%$); $12\,276$ non-finite exclusions counted separately; $2\,025$ of
$18\,000$ generations ran the weighted centroid with finite candidates
excluded; $1\,016$ generations fell back to the population mean, none of them
caused by non-finite states alone; mean eligible set $35.56$ of $40$, minimum
$0$, maximum $40$.

\paragraph{Sensitivity design.} The variant core differs from the frozen core
in exactly two places: the function name, and \verb|vd=fitness<1e7| becoming
\verb|vd=fitness<Inf|. The build reverses both edits and requires byte-exact
reproduction of the frozen source before the variant is written, so no other
change can be present. Seeds, initial populations, population size, iteration
count and the 1040-evaluation budget are unchanged, and the run reuses the
archived seed map. Each run was additionally required to return $1040$
evaluations, $25$ iterations, an FE history of length $1040$ and a terminal FE
history value equal to the reported best.

\paragraph{Negative controls.} The $540$ runs in which the frozen audit had
recorded zero finite activation were run through the variant and all $540$
reproduced the frozen terminal value exactly. The intervention therefore
changes behaviour only where the threshold was mechanically active, which is
what makes the $180$ of $720$ active runs interpretable as a controlled
contrast.

\paragraph{Per-cell sensitivity results.} Thirty matched seeds per cell. The
paired log shift is the median of the per-seed $\log_{10}$ differences,
positive meaning the no-cutoff variant has the larger error; $r_{\mathrm{rb}}$
is the matched-pairs rank-biserial correlation; $p$ is a Wilcoxon signed-rank
test on the untransformed paired differences, Holm-corrected across the six
cells.

\begin{center}\footnotesize
\begin{tabular}{lrrrcc}
\hline
Cell & Paired log shift & $r_{\mathrm{rb}}$ & Raw $p$ & Holm sig. & No-cut/tie/frozen\\
\hline
F1, $D{=}10$ & $-0.000$ & $-0.032$ & $0.883$ & no & 17/0/13\\
F6, $D{=}10$ & $+0.044$ & $+0.080$ & $0.715$ & no & 13/0/17\\
F8, $D{=}10$ & $\;\;0.000$ & $+0.124$ & $0.582$ & no & 14/3/13\\
F1, $D{=}20$ & $+0.056$ & $-0.389$ & $0.064$ & no & 10/0/20\\
F6, $D{=}20$ & $+0.141$ & $-0.256$ & $0.229$ & no & 12/0/18\\
F8, $D{=}20$ & $-0.000$ & $-0.090$ & $0.686$ & no & 16/1/13\\
\hline
\end{tabular}
\end{center}

\noindent Across all $180$ active runs the no-cutoff variant was better, tied
or worse in $82/4/94$ seed pairs. For F6 at $D=10$ the independently computed
medians fall from $3.22\times10^{4}$ to $1.65\times10^{4}$, which would suggest
a large improvement, while the median paired shift is $+0.044$ decades with the
variant better on only $13$ of $30$ seeds. The quantities differ because
$\log_{10}\mathrm{med}(V)-\log_{10}\mathrm{med}(F)$ is not
$\mathrm{med}[\log_{10}V_i-\log_{10}F_i]$; the matched-seed design makes the
latter the one that supports inference.

\paragraph{Paired analysis.} Errors are floored at $10^{-8}$ before
taking logarithms, the same convention used elsewhere in this work. The
paired statistic is $\mathrm{med}[\log_{10}V_i-\log_{10}F_i]$ over the thirty
matched seeds, tested with a Wilcoxon signed-rank test on the untransformed
paired differences and corrected across the six cells by Holm's method. 

\paragraph{Scope.} The $f_{\mathrm{ctr}}=\infty$ run is a post-campaign
sensitivity variant and a mechanism validation, not a replacement optimizer
campaign and not a redesign of REMBGO. A rank- or quantile-based eligibility
rule would remove the scale dependence and is identified here only as a possible
future refinement. No result reported elsewhere in this paper was produced
by the variant, and the frozen benchmark results are unchanged.